\pdfoutput=1

\documentclass[11pt]{article}

\usepackage[table]{xcolor}
\usepackage{etoolbox}

\usepackage{supertabular}
\usepackage{longtable}
\usepackage{setspace}
\usepackage{placeins}
\usepackage{stfloats}
\usepackage{mdframed}
\usepackage[]{acl}

\usepackage{fancyvrb}
\usepackage{pythonhighlight}
\usepackage{times}
\usepackage{latexsym}
\usepackage{adjustbox}
\usepackage{xcolor}
\definecolor{mygray}{gray}{0.9}
\usepackage[T1]{fontenc}

\usepackage[utf8]{inputenc}

\usepackage{microtype}

\usepackage{inconsolata}
\usepackage{graphicx}
\usepackage{subcaption}
\usepackage{booktabs}
\usepackage{adjustbox}
\usepackage{multirow}
\usepackage{svg}
\usepackage{xcolor}

\usepackage{pgf} %
\definecolor{low}{HTML}{2e87ec}  %
\definecolor{high}{HTML}{ec462e}  %
\newcommand*{\opacity}{60}%
\newcommand*{\minval}{0.0}%
\newcommand*{\maxval}{1.0}%
\newcommand{\gradient}[1]{
    \ifdimcomp{#1pt}{>}{\maxval pt}{#1}{
        \ifdimcomp{#1pt}{<}{\minval pt}{#1}{
            \pgfmathparse{int(round(100*(#1/(\maxval-\minval))-(\minval*(100/(\maxval-\minval)))))}
            \xdef\tempa{\pgfmathresult}
            \cellcolor{high!\tempa!low!\opacity} #1
    }}
}

\usepackage[capitalize, nameinlink]{cleveref}
\crefformat{section}{Section #2#1#3} %
\crefformat{subsection}{Section #2#1#3}
\crefformat{subsubsection}{Section #2#1#3}
\crefformat{figure}{Figure #2#1#3}
\crefformat{appendix}{App. #2#1#3}

\newenvironment{itemize*}%
  {\begin{itemize}%
    \setlength{\itemsep}{0pt}%
    \setlength{\parskip}{0pt}%
    \setlength{\topsep}{0pt}}%
  {\end{itemize}}

\usepackage[nameinlink]{cleveref}

\crefformat{section}{\S#2#1#3}
\crefformat{subsection}{\S#2#1#3}
\crefformat{subsubsection}{\S#2#1#3}
\title{AI `News' Content Farms Are Easy to Make and Hard to Detect: \\ A Case Study in Italian}

\author{
  {\bf Giovanni Puccetti\textsuperscript{$\alpha$}, Anna Rogers\textsuperscript{$\beta$}, Chiara Alzetta\textsuperscript{$\gamma$}, Felice Dell'Orletta\textsuperscript{$\gamma$}, Andrea Esuli\textsuperscript{$\alpha$}}\\
  \textsuperscript{$\alpha$} Istituto di Scienza e Tecnologia dell'Informazione ``A. Faedo'' \\ \texttt{\{giovanni.puccetti,andrea.esuli\}@isti.cnr.it} \\ \textsuperscript{$\beta$} IT University of Copenhagen \\ \texttt{arog@itu.dk} \\ \textsuperscript{$\gamma$} ItaliaNLP Lab, Istituto di Linguistica Computazionale ``Antonio Zampolli'' \\ \texttt{\{chiara.alzetta,felice.dellorletta\}@ilc.cnr.it}
}

\newcommand{\lscit}{\emph{llama-2-7b\_it}}
\newcommand{\lmcit}{\emph{llama-2-13b\_it}}
\newcommand{\mcit}{\emph{mistral\_it}}
\newcommand{\lscits}{\emph{llama-2-7b\_it\_3981}}
\newcommand{\lmcits}{\emph{llama-2-13b\_it\_3981}}
\newcommand{\mcits}{\emph{mistral\_it\_3981}}
\newcommand{\lscitm}{\emph{llama-2-7b\_it\_7862}}
\newcommand{\lmcitm}{\emph{llama-2-13b\_it\_7862}}
\newcommand{\mcitm}{\emph{mistral\_it\_7862}}

\usepackage{todonotes}

\begin{document}
\maketitle

\begin{abstract}
Large Language Models (LLMs) are increasingly used as `content farm' models (CFMs), to generate synthetic text that could pass for real news articles. This is already happening even for languages that do not have high-quality monolingual LLMs. We show that fine-tuning Llama (v1), mostly trained on English, on as little as 40K Italian news articles, is sufficient for producing news-like texts that native speakers of Italian struggle to identify as synthetic.

We investigate three LLMs and three methods of detecting synthetic texts (log-likelihood, DetectGPT, and supervised classification), finding that they all perform better than human raters, but they are all impractical in the real world (requiring either access to token likelihood information or a large dataset of CFM texts). We also explore the possibility of creating a proxy CFM: an LLM fine-tuned on a similar dataset to one used by the real `content farm'. We find that even a small amount of fine-tuning data suffices for creating a successful detector, but we need to know which base LLM is used, which is a major challenge.

Our results suggest that there are currently no practical methods for detecting synthetic news-like texts `in the wild', while generating them is too easy. We highlight the urgency of more NLP research on this problem.
\end{abstract}

\maketitle

\section{Introduction}
\label{sec:intro}
\begin{figure}[t]
    \centering
    \includegraphics[width=\linewidth,keepaspectratio]{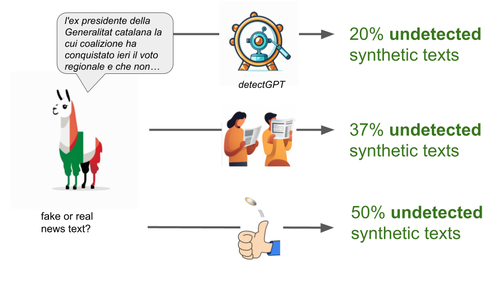}
    \caption{Detecting synthetic Italian news text generated by fine-tuned Llama-65B: error rates for DetectGPT, native speakers of Italian and random guess.}
    \label{fig:graphical_abstract}
\end{figure}

The modern Large Language Models (LLMs) can generate increasingly fluent and plausible-sounding texts, which sparks concerns about their potential misuse by bad actors.
One of the emerging problems is AI-driven news ``content farms'': news-like sites filled with synthetic texts that are not necessarily serving a misinformation campaign, but are plausible-looking enough to deceive the readers and generate web traffic. Already in May 2023 NewsGuard reported that they found 49 such `outlets' \cite{SadeghiArvanitis_2023_Rise_of_Newsbots_AI-Generated_News_Websites_Proliferating_Online}, and on June 4th 2024 their count\footnote{\url{https://www.newsguardtech.com/special-reports/ai-tracking-center/}} was 840. Sometimes such sites publish original `content', and sometimes they automatically `rewrite' articles from real news outlets without attribution \cite{BrewsterWangEtAl_2023_Plagiarism-Bot_How_Low-Quality_Websites_Are_Using_AI_to_Deceptively_Rewrite_Content_from_Mainstream_News_Outlets_Misinformation_Monitor_August_2023}.
They are primarily created for serving programmatic ads, and even major brands may unwittingly support such `outlets' \cite{Ryan-Mosley_2023_Junk_websites_filled_with_AI-generated_text_are_pulling_in_money_from_programmatic_ads}.
They already operate in many languages.\footnote{NewsGuard AI Tracking Center currently states that they found such `outlets' in Arabic, Chinese, Czech, Dutch, English, French, German, Indonesian, Italian, Korean, Portuguese, Russian, Spanish, Tagalog, Thai, and Turkish.}  %
The problem will likely only get worse with time, and it needs more attention both on the policy \& regulation side and from the NLP researchers.

We illustrate how easy it is to create `content farm' models (CFMs), and how hard to detect them, by considering a case that should be relatively tricky. In \cref{sec:llama} we successfully turn Llama, a relatively old LLM mostly trained on English, into an Italian news CFM -- by fine-tuning it on only 40k Italian news texts.
For Llama65B this turns out to be sufficient to mislead native speakers of Italian, who identify synthetic texts with only 64\% accuracy, vs 50\% random guess (see \cref{sec:human}). %
We also find that existing detection methods perform better than humans (\cref{sec:auto}), but are impractical in the real world (\cref{sec:practicability}). %

In \cref{sec:fine_tuning_detection} we explore an alternative approach: fine-tuning another LLM as a proxy for the real CFM and relying on its token likelihood scores as proxies for the scores of the real CFM. We find that this works with even a small amount of fine-tuning data (only 3\% of the full fine-tuning dataset) but only given that we know which base LLM was used as CFM. We also experiment with using these proxy scores to identify the base LLM, but this method would also be impractical if there are dozens, if not hundreds possible alternatives.

We hope that our findings will boost similar investigations for other languages, and highlight the urgency of developing model-agnostic methods for synthetic text detection. To facilitate future research on Italian, we release \textit{(i)} 15k news passages generated by models fine-tuned on the CHANGE-it dataset, an Italian news dataset \cite{changeit_2020}, \textit{(ii)} ratings produced by 5 human annotators on 400 texts, with a balanced distribution of 50\% human-written and 50\% synthetic passages, and \textit{(iii)} 600k synthetic alterations of both the original samples from the CHANGE-it dataset and the synthetic texts. The code and data are publicly available.\footnote{\url{https://github.com/gpucce/synthetic\_llm\_data}}

We will not publicly release our fine-tuned LLMs (since they are best suited to be used as CFMs), but we welcome direct requests from researchers working on this problem.

\section{Related work}

Driven by the increasing number of strong openly available LLMs \cite{touvron2023llama, brown_2020_gpt3, raffel_2020_t5,jiang2023mistral},
several studies focused on the detection of synthetic text detection.

\citet{ghosal2023a} identified two main groups of approaches: those based on token likelihood, and supervised classification. Among the former, \citet{mitchell_detectgpt_2023, su-etal-2023-detectllm, hans2024spotting, gehrmann-etal-2019-gltr, mireshghallah-etal-2024-smaller} proposed detection methods that rely on language models token distribution. For the supervised detection, \citet{verma2023ghostbuster} proposed an %
approach that relies on the availability of labelled datasets of synthetic and human-written texts.

\citet{chakraborty-etal-2023-counter} propose a 6-way split of synthetic text detection methodologies: (i) watermarking, (ii)
perplexity estimation, (iii) burstiness estimation, (iv) negative log-likelihood curvature, (v) stylometric variation, and (vi) classifier-based approaches.

After the release of ChatGPT \cite{OpenAI_2022_Introducing_ChatGPT} and GPT-4 \citep{OpenAI_2023_GPT-4_Technical_Report}, a lot of studies (mostly not yet peer reviewed) focused specifically on the detection of text generated by these models \cite{DhainiPoelmanEtAl_2023_Detecting_ChatGPT_Survey_of_State_of_Detecting_ChatGPT-Generated_Text}. Since it is not possible to access token probabilities for a candidate generated text, some of this work relies on a proxy model \citep{VasilatosAlamEtAl_2023_HowkGPT_Investigating_Detection_of_ChatGPT-generated_University_Student_Homework_through_Context-Aware_Perplexity_Analysis}, but most rely on supervised classification \citep{MitrovicAndreolettiEtAl_2023_ChatGPT_or_Human_Detect_and_Explain_Explaining_Decisions_of_Machine_Learning_Model_for_Detecting_Short_ChatGPT-generated_Text,LiaoLiuEtAl_2023_Differentiating_ChatGPT-Generated_and_Human-Written_Medical_Texts_Quantitative_Study,GuoZhangEtAl_2023_How_Close_is_ChatGPT_to_Human_Experts_Comparison_Corpus_Evaluation_and_Detection,LiuZhangEtAl_2023_ArguGPT_evaluating_understanding_and_identifying_argumentative_essays_generated_by_GPT_models}, including OpenAI itself \citep{KirchnerAhmadEtAl_2023_New_AI_classifier_for_indicating_AI-written_text}. Sometimes classical machine learning techniques are reported to perform well: e.g. \citet{DesaireChuaEtAl_2023_Accurately_detecting_AI_text_when_ChatGPT_is_told_to_write_like_chemist} report a high performance on chemistry articles with XGBoost classifier and 20 features extracted from paragraphs.

Recently, several benchmarks for the synthetic text detection task have been released, \citet{wang2023m4,wang2024m4gtbench} contribute a large multilingual, multi-domain dataset that can be used to benchmark synthetic text detection systems.
\citet{macko-etal-2023-multitude} propose a benchmark for the detection of text generated by multilingual LLMs, while \citet{dugan2024raid} propose a very large benchmark for synthetic text detection, controlling for the temperature used to
generated the text, which has been shown to be relevant for detection \cite{mitchell_detectgpt_2023}.

The above works focus on the scenario where the effort of detecting synthetic texts is on the user side.
The complementary direction on the developer side is watermarking: ensuring that the LLM output creates some kind of statistical ``signature'' that would help to identify it.
There are multiple proposals for how to do this \cite[inter alia]{FernandezChaffinEtAl_2023_Three_Bricks_to_Consolidate_Watermarks_for_Large_Language_Models,KirchnerAhmadEtAl_2023_New_AI_classifier_for_indicating_AI-written_text,KuditipudiThickstunEtAl_2023_Robust_Distortion-free_Watermarks_for_Language_Models,LiChengEtAl_2023_PLMmark_Secure_and_Robust_Black-Box_Watermarking_Framework_for_Pre-trained_Language_Models,TakezawaSatoEtAl_2023_Necessary_and_Sufficient_Watermark_for_Large_Language_Models,WuHuEtAl_2023_DiPmark_Stealthy_Efficient_and_Resilient_Watermark_for_Large_Language_Models,YooAhnEtAl_2023_Advancing_Beyond_Identification_Multi-bit_Watermark_for_Large_Language_Models}, and some initial results suggesting that such techniques could be sufficiently robust to human and machine paraphrasing \cite{pmlr-v202-kirchenbauer23a,kirchenbauer2024on}. However, most of the current `open' LLMs are freely available without any watermarking, and as we will show, they are already sufficient to be used as CFMs.

For the problem of detecting synthetic text without watermarking, the current research focuses on either monolingual or multilingual-by-design LLMs, and most studies do not focus on a specific domain. We stress the importance of also investigating fine-tuned models because LLMs are expensive to both train and run inference on \cite{samsi2023words}. Hence, starting from a public, relatively small, but high-quality LLM and fine-tuning it for a specific type of content is probably the most plausible scenario for `content farms' that aim to produce texts cheaply. To the best of our knowledge, this is the first study to focus on the scenario where the `CFM' is fine-tuned for news, and in a language that it was not meant for originally. Moreover, except for the recent work by \citet{wang2024m4gtbench}, existing resources do not cover Italian.

\section{Fine-tuning of Llama as an Italian `Content Farm' Model}
\label{sec:llama}

As our `CFM', we choose the original Llama model \citep{touvron2023llama}, in
7B and 65B parameter versions. Since this is one of the first public high-performing LLMs of this size, our Llama results serve as a lower bound for what could be expected from later LLMs, such as Llama 2 \cite{touvron2023llama2} and Llama 3,\footnote{\url{https://ai.meta.com/blog/meta-llama-3/}} Aya \cite{aryabumi2024aya}, and others. We do \textit{not} suggest that it is possible to obtain good results with any language and any LLM (see also \cref{sec:limitations}): such a transfer depends on the similarity between languages and their coverage in the training data. But given that `content farms' have already been identified in at least 16 languages (see \cref{sec:intro}), many others could follow.

Our choice of Italian enables
a lower-bound estimate of what could be expected of monolingual or multilingual LLMs with more exposure to the target language. Llama is a `mostly-English' model, not intended to be multilingual. The resources it was trained on, such as C4 corpus \cite{raffel_2020_t5}, made deliberate efforts to filter out non-English text. However, it was exposed to at least Italian Wikipedia \citep[][p.2]{touvron2023llama}. Hence, Wikipedia is likely the main, if not the only source of Italian in Llama.

We experiment with the original Llama baseline (65B pre-trained model with no extra training), and two versions of our Llama fine-tuned on Italian news, after 20K and 60K steps. %
The technical details for fine-tuning are provided in \cref{app:finetuningllama1}.
We remark that creating such a CFM now comes with very few technical or financial difficulties,\footnote{Renting GPUs on cloud providers such as Amazon is now relatively cheap (approx 15\$ per hour for 8x40Gb A100) and would require as little as 100\$ to replicate one of our LLM training sessions and data generation. The technical barrier to fine-tuning LLMs is also low now, thanks to tools like Huggingface's Autotrain:\url{https://huggingface.co/autotrain}.
We do not criticize open-sourcing such tools, but we hope that our results would highlight the necessity of more research on synthetic text detection.} which could increase the number of bad actors.

\section{Detection of Synthetic News in Italian By Native Speakers}
\label{sec:human}

\paragraph{Methodology.} To assess whether native Italian speakers would be able to identify synthetic news texts generated by our CFM, we set up a crowd-based study following the general recommendations for human evaluation of automatically generated texts proposed by \citet{VANDERLEE2021101151}. Specifically, we created 4 surveys, with 100 questions in each.  To maintain the rater engagement,  we administered the surveys in five sessions, each comprising 20 questions. The raters were anonymously recruited among Italian native speakers via the Prolific\footnote{\url{https://www.prolific.co/}. 
The participant group was balanced in terms of gender (49.46\% female) and student status (50.60\% reported being students).} online crowd-sourcing platform. 
Five different raters participated in each survey session, with no limitations on the number of sessions a rater could undertake.
The 20-question sessions took 8 mins 23 secs on average, and the raters were compensated at 9,68\$ per hour.\footnote{This hourly payment rate was certified as `Fair' by the Prolific platform.}
The study involved a total of 93 different raters, with an average age of 32.01 ($\pm$10.76). %

Each of the 4 surveys is designed to assess the texts generated by a different model: Llama 7B and 65B, both with and without fine-tuning on Italian news. Each question required raters to read two texts, denoted as $A$ and $B$, and answer the question \emph{``Text B follows text A, do you think text B is written by a machine?''}. The answer was a rating on a 5-point Likert scale, where 1 indicates `certainly human-written' and 5 `certainly machine-generated'. In 50\% of the questions, both text $A$ and $B$ were human-written news articles coming from the original CHANGE-it test dataset. %

The topics included daily national political events (e.g. politicians' declarations), general news (e.g. climate catastrophes), and relevant international news (e.g. European leadership meetings).

To estimate the accuracy of human raters on this task, we map the scores on the 5-point scale to a binary score by computing the average score assigned by the raters to a given sample.
We interpret the average score above 3 as indicating that a given sample $B$ was generally perceived as machine-generated. We also experimented with a different thresholding approach, using the average rating as a threshold, that showed similar results; these results are available in \cref{app:human_eval_scaled_mean}.

\begin{table}[t]
\footnotesize
    \centering
\begin{tabular}{cccc}
\toprule
Model & Accuracy & STD & Fleiss k\\
\midrule
\emph{Llama 7B} pretrain & 83.2 & 7.0 & 0.36 \\
\emph{Llama 7B} finetuned & 69.5 & 12.2 & 0.22 \\
\midrule
\emph{Llama 65B} pretrain & 73.7 & 5.8 & 0.33 \\
\emph{Llama 65B} finetuned & 64.2 & 11.2 & 0.21 \\
\bottomrule
\end{tabular}%
    \caption{Accuracy and standard deviation achieved by human raters in assessing human-written versus machine-generated news. We report the inter-rater agreement measured as group Fleiss' k.}
    \label{tab:human_eval_mean}
\end{table}

\paragraph{Results.} \Cref{tab:human_eval_mean} shows the outcome of our analysis. Since the raters' accuracy in detecting news generated by the largest fine-tuned Llama 65B is as low as 64\%, we can answer our research question positively: \textbf{Llama can be fine-tuned to generate hard-to-detect news-like text in Italian}. We only used 40K samples for fine-tuning a relatively old model, so even more plausible-sounding synthetic text could likely be created with more data and more recent LLMs. %

Overall, the raters' accuracy reflects two foreseeable
trends: the smaller 7B models are easier to detect, and the fine-tuned models are harder to detect. %
Interestingly, the small 7B version, when fine-tuned on Italian, is identified by raters with accuracy close to the larger pretrained 65B.

\Cref{tab:human_eval_mean} also reports the average inter-rater agreement for each survey. Fleiss $\kappa$ \citep{fleiss1971mns} is in the range between 0.21-0.36, indicating a ``fair'' agreement \citep{landis1977measurement} consistent with similar human-evaluation studies \citep{VANDERLEE2021101151}. The raters agree more strongly when assessing non-fine-tuned models, possibly because they occasionally switch from Italian to English mid-generation, a characteristic identified by raters as indicative of machine-generated texts. \autoref{fig:it-en-switch} shows an example of such a switch. %

\begin{figure}[t]
\begin{mdframed}[backgroundcolor=mygray,linewidth=0pt]
\footnotesize
\textbf{Prompt:} \emph{"[...] l'ex presidente della Generalitat catalana la cui coalizione ha conquistato ieri il voto regionale e che non..."}
{\scriptsize \color{gray}
EN: [...] the former president of Catalan Generalitat, whose coalition won the regional election yesterday, and who...}

\textbf{Pre-trained:} \emph{... vuole rinunciare alla secessione. \underline{In the 6-week period prior to 12/06/19}...}
{\scriptsize \color{gray}
EN: ... does not want to give up the secession. In the 6-week period prior to 12/06/19 ...}

\textbf{Fine-tuned:} \emph{...aveva perso tempo per dire la sua. Da Bruxelles, dove si trova da allora} ...
{\scriptsize \color{gray}
EN: ... does not waste time to mention his opinion. From Brussels, where he resides since...}

\end{mdframed}
\caption{Example: without fine-tuning on Italian, Llama is prone to switching to English.}
\label{fig:it-en-switch}
\end{figure}

The qualitative analysis of 100 machine-generated instances (25 per model) showed that
46/100 examples had no obvious issues with language, but, as expected for LLM-generated texts, their content was factually incorrect and, worryingly, the factual errors were not necessarily obvious without extra fact-checking effort. %
Additionally, in this sample of 100 texts, we found 7 examples where the generated text contradicted the prompt, 8 cases of language-switching, 18 samples with grammatical errors and 21 with expressions that are grammatically correct but unnatural in Italian. %
Annotated examples for each model are shown in \cref{app:examples}. %

We stress that this study focuses only on the problem of synthetic news-like text, which is sufficiently plausible for the `content farms' to lead the users to `news' websites and be served ads. Their success also likely depends on the quality of the `headlines', their match to the interests of the audience, the position in the search engine rankings and other factors beyond the scope of this study. Still, having such texts is a necessary, though not sufficient condition for operating a `content farm'.

\begin{figure*}[ht]
    \centering
    \begin{subfigure}{0.31\textwidth}
        \includegraphics[width=\linewidth]{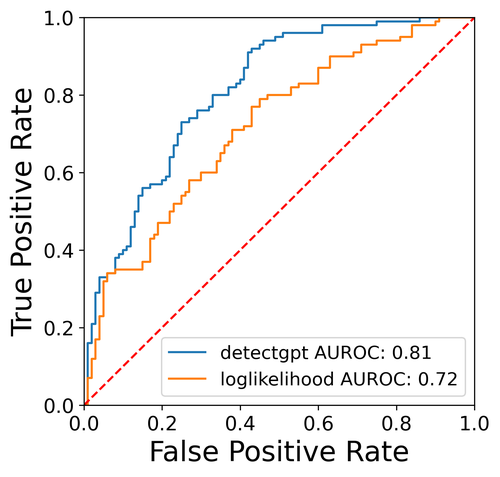}
        \subcaption{Original Llama 65B}
        \label{subfig:roc_pretrain_ita}
    \end{subfigure}
    \begin{subfigure}{0.31\textwidth}
        \includegraphics[width=\linewidth]{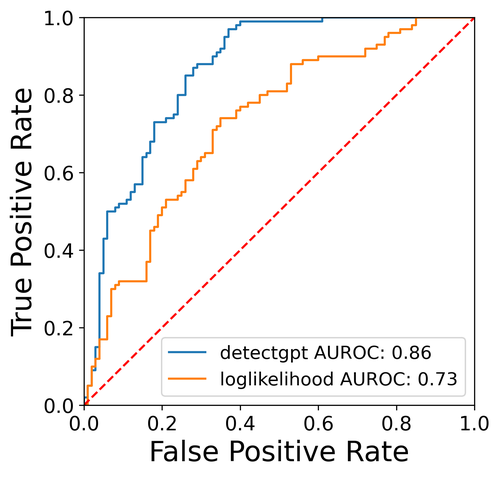}
        \subcaption{Fine-tuned epoch 2}
        \label{subfig:roc_fine_tune_e2}
    \end{subfigure}
    \begin{subfigure}{0.31\textwidth}
        \includegraphics[width=\linewidth]{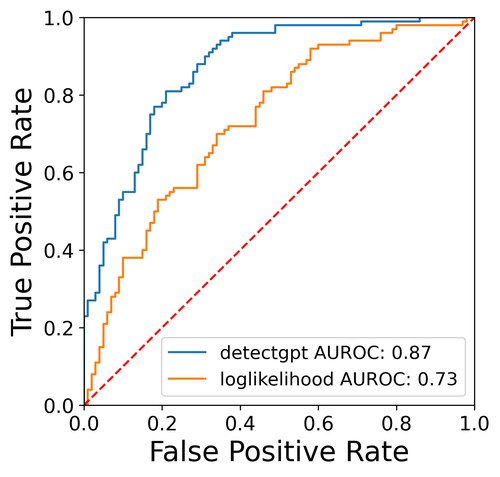}
        \subcaption{Fine-tuned epoch 6}
        \label{subfig:roc_fine_tune_e6}
    \end{subfigure}
    \caption{ROC curve for \emph{DetectGPT} and \emph{log-likelihood}. In
    (\subref{subfig:roc_pretrain_ita}) for Llama 65B measured over 100 sentences from the CHANGE-it data-set (Italian), in (\subref{subfig:roc_fine_tune_e2}) the same measure for Llama 65B model after 20,000 fine tuning steps on CHANGE-it training set and in (\subref{subfig:roc_fine_tune_e6}) after 60,000 fine-tuning steps.}
    \label{fig:auroc_ita}
\end{figure*}

\section{Automatic Detection of Synthetic News in Italian}
\label{sec:auto}

\subsection{Approaches based on token likelihood}
\label{sec:log}

\paragraph{Methodology.} Similarly to \citet{jawahar-etal-2020-automatic, sadasivan_can_2023, chakraborty_possibilities_2023}, we attempt the zero-shot detection of artificially generated text. We experiment with two approaches for synthetic text detection. Both assume to have access to the likelihood of each token in a sentence, according to the model whose ``authorship'' is under analysis.

The \emph{log-likelihood} measures how likely a sentence is according to the probability assigned by the model to each token.
The core idea behind the \emph{DetectGPT} score \cite{mitchell_detectgpt_2023} is getting a more robust score by normalising the \emph{log-likelihood} of a sentence based on modifications\footnote{In our case, we are interested in the likelihood of a synthetic sentence estimated by Llama, vs the normalized likelihood that Llama assigns to modifications of that same sentence that we generate with T5 \cite{raffel_2020_t5} as the bootstrap model. See \cref{app:examples} for examples of such modifications.} of that same sentence generated by a different model (which we refer to as the \textit{bootstrap model}).

Given a sentence, both \emph{log-likelihood} and \emph{DetectGPT} can be used with a threshold to tell if that sentence is more likely written by a human or by a language model. The threshold is estimated empirically based on a collection of synthetic and real samples. We refer the reader to \citet{mitchell_detectgpt_2023} for details of likelihood-based methods.

We start by validating our codebase and implementation by repeating prior experiments on xsum \cite{narayan_2018_xsum} dataset in English, and we obtain comparable results (see \cref{app:xsum_log_experiments}).

To measure the ROC of both \emph{DetectGPT} and \emph{log-likelihood} we select 100 random sentences and generate\footnote{All generation tasks are performed using Nucleus Sampling for decoding \cite{Holtzman2020The}.} another 100 by %
using the first 30 tokens of each sentence as a prompt for the model under analysis. After this, we clip all sentences to 150 tokens,\footnote{Due to different tokenizers, this step results in sentences with a varying number of words, but a similar length.} measure the score for each sample, and compute the AUROC on all 200 sentences (half human-written and half machine-completed).

To compare the detection approaches for the original and fine-tuned Llama, we use the CHANGE-it test dataset for Italian.
As the bootstrap model, we use \emph{IT5-large} \citep{sarti2022it5} for Italian.

\paragraph{Results.}

\Cref{fig:auroc_ita} shows the AUROC of \emph{DetectGPT} and \emph{log-likelihood} for all our models. \emph{log-likelihood} %
does not seem to react to fine-tuning at all: the fine-tuned models have almost the same AUROC as the pre-trained one. \emph{DetectGPT} does have a 5-6 points higher AUROC for the fine-tuned models. %
However, by qualitative analysis, see \cref{sec:human}, we find that fine-tuning should have made the task more rather than less difficult. Without fine-tuning, our Llama CFM has a tendency to switch to English mid-generation, as we also observed this in the human detection study (see \autoref{fig:it-en-switch}, more examples available in \cref{app:examples}). This language switch should be a clear marker for detecting both the original 7B and the 65B models, and it vanishes after fine-tuning. But both log-likelihood and DetectGPT are missing this clear signal. %

Although the \textit{DetectGPT} and \textit{log-likelihood} perform relatively well in our tests, we stress that this indicates a measure of the difficulty of this task, rather than a solution to synthetic news detection (see \autoref{sec:practicability}). We remark that DetectGPT score can be turned into an accuracy measurement by fixing a threshold, for a direct comparison with the human evaluation accuracy. In our case, DetectGPT accuracy is $\approx 80\%$ using the median\footnote{We choose this threshold knowing that the dataset is balanced and that DetectGPT is monotonic, otherwise we would need to tune it.} score as the threshold for fine-tuned Llama 65B.

\subsection{Supervised Detection of Synthetic Texts}
\label{sec:supervised_detection}

A different approach to identifying synthetic texts is to train supervised classifiers \cite{liu2019roberta, conneau-etal-2020-unsupervised}.
However, this requires a labelled and balanced dataset of human and synthetic texts. To understand the challenges of this scenario, we use DICE, a different Italian news dataset focusing on crime news \cite{bonisoli2023}. We mix DICE with the CHANGE-it data as well as the synthetic texts. This simulates a more realistic data collection process compared to previous studies that solely focused on fully in-domain data. %

\paragraph{Methodology.}

To create synthetic news, we fine-tune the following recent models: llama2-7b, llama2-13b \cite{touvron2023llama2} and Mistral-7b \cite{jiang2023mistral} on the full CHANGE-it training set (see \cref{app:finetuningllama2} for the fine-tuning details).
We refer to the resulting fine-tuned models as \lscit, \lmcit{}, and \mcit{}.

For each of these models, we create a suite of datasets with training sets composed of 2K, 4K or 8K samples. In all settings, the test sets comprise 2K samples, namely 1K texts from the CHANGE-it test set and 1K synthetic news with the same titles.
The training sets are built in two ways.
In \emph{in domain} setting, 50\% texts are synthetic, and 50\% are articles sourced from CHANGE-it. In \emph{mixed source}, the human-authored articles come from two different datasets, 25\% each (CHANGE-it and DICE). \emph{Mixed source} is closer to the scenario where we do not know what dataset was used to fine-tune the CFM, and sample from a wide range of possible news articles. The more diverse this set of non-synthetic examples, the harder the classification task will probably become.

\begin{figure}[t]
    \centering
    \includegraphics[width=\linewidth]{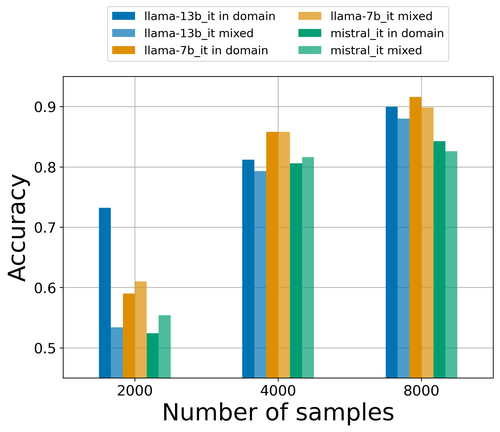}
    \caption{Accuracy of classifier based on xlm-RoBERTa-large for human/synthetic text classification task, for synthetic texts generated by three LLMs fine-tuned on CHANGE-it. The classifier was trained on 50\% synthetic texts and either 50\% CHANGE-it texts (\emph{in domain}), or 25\% texts from CHANGE-it and 25\% from DICE (\emph{mixed source}). Classification is only successful at at least 4K labeled samples, and the \emph{mixed source} scenario is consistently more challenging.}
    \label{fig:supervised_detection_xlm}
\end{figure}

Similarly to \citet{wang2023m4}, we experiment with two classifiers, based on RoBERTa \cite{liu2019roberta} and XLM-RoBERTa \cite{conneau-etal-2020-unsupervised}.
The former is pre-trained on English and the latter specifically made for multilingual settings. We use the same hyper-parameters for both classifiers, and we train them for 3 epochs. The technical details are provided in \cref{app:finetuningclassifier}.

\paragraph{Results.} \Cref{fig:supervised_detection_xlm} shows that at 8k training samples \emph{xlm-roberta-large} accuracy ranges between 84\% and 92\% depending on the generator.
With 4k training samples, the accuracy decreases between
81\% and 86\%, with a similar difference between \emph{in domain} and \emph{mixed source}. With 2k training samples, the performance for all but 3 models drops to under 56\% -- i.e. almost random chance level. %
The RoBERTa-based classifier yields a similar trend. See \cref{app:finetuningclassifier} for the graph for RoBERTa and numerical results for both experiments.

We find that for these datasets there is a threshold between 2k and 4k samples that marks a strong difference in the performance of the supervised classifier (20-25\% loss in accuracy), while increasing the dataset size further provides diminishing returns (2-8\% gain). Our \emph{mixed source} scenario at 8K samples is also consistently harder for the classifier; at 4K the results are mixed across different LLMs, and at 2K it is hard to draw firm conclusion for \lscit{} and \mcit{} because their performance is generally low. This is with a multilingual classifier base; for RoBERTa-based classifier, the \emph{mixed} setting is consistently harder in all settings (see \cref{app:finetuningclassifier}). %

\section{Detecting CFMs with Proxy Models}
\label{sec:fine_tuning_detection}

\subsection{How Much Fine-tuning Do We Need?}

So far we only considered the scenarios where the model generating synthetic texts is the same that computes the likelihood used to detect them. %
Let us now consider an alternative: a \textit{proxy model} approximating the likelihood scores of the CFM. This could be expected to work if the proxy model and the CFM were fine-tuned on the same dataset. But %
that assumption is also too strong to be practical.

We explore relaxing it further to the scenario where we only have access to a small set of texts similar to the fine-tuned model outputs. This could be a small subset of the fine-tuning dataset or, more generally, samples from a similar distribution (e.g. different samples from a given newspaper or a social media account).
We ask whether this is enough: \textbf{can we train a model on a small set of in-distribution texts, so that the likelihood it assigns to tokens is sufficient to detect texts from a fully fine-tuned CFM?}

\paragraph{Methodology.}

To answer this question, we fine-tune the three LLMs used in the supervised detection experiment in \cref{sec:supervised_detection} (namely  \lscit{}, \lmcit{} and \mcit{}) on varying subsets of the CHANGE-it dataset. We then use their likelihood and DetectGPT score to detect synthetic text from models fine-tuned on the full CHANGE-it dataset, as described in \cref{sec:auto}.

For that, we select 5000 samples from the CHANGE-it test set, and for each sample, we prompt all models with the title and initial tokens of the original article (see \autoref{app:prompting} for more details). %
This results in 3 datasets with 10k samples in each, 5k human written (from the CHANGE-it test set) and 5k synthetically generated news with the same titles%
, which can be used as a benchmark for the detection of synthetic news in Italian.

Afterwards, we select two subsets of the CHANGE-it training dataset with 3981 and 7862 samples, respectively 3\% and 6\% times the original fine-tuning dataset size.
We fine-tune the same LLMs on both of these smaller training sets, see \cref{app:finetuningllama2} for details on the fine-tuning procedure and \cref{tab:model_naming} for a summary of the model naming.

\begin{table*}
\footnotesize
\centering
\begin{tabular}{rcccccc}
\toprule
Detector model & \multicolumn{6}{c}{Generator model}\\
 \cmidrule{2-7}
 & \multicolumn{2}{c}{\lmcit{}} & \multicolumn{2}{c}{\lscit} & \multicolumn{2}{c}{\mcit{}} \\

 & dGPT & llh & dGPT & llh & dGPT & llh \\
\midrule
\emph{llama-2-13b} & \gradient{0.73} & \gradient{0.61} & \gradient{0.54} & \gradient{0.40} & \gradient{0.56} & \gradient{0.43} \\
\lmcits{} & \gradient{0.84} & \gradient{0.69} & \gradient{0.53} & \gradient{0.35} & \gradient{0.56} & \gradient{0.42} \\
\lmcitm{} & \gradient{0.85} & \gradient{0.70} & \gradient{0.53} & \gradient{0.34} & \gradient{0.56} & \gradient{0.41} \\
\lmcit{} & \gradient{0.87} & \gradient{0.70} & \gradient{0.48} & \gradient{0.27} & \gradient{0.55} & \gradient{0.39} \\
\midrule
\emph{llama-2-7b} & \gradient{0.58} & \gradient{0.49} & \gradient{0.75} & \gradient{0.59} & \gradient{0.57} & \gradient{0.46} \\
\lscits{}       & \gradient{0.63} & \gradient{0.48} & \gradient{0.86} & \gradient{0.67} & \gradient{0.60} & \gradient{0.45} \\
\lscitm{}       & \gradient{0.63} & \gradient{0.47} & \gradient{0.87} & \gradient{0.68} & \gradient{0.60} & \gradient{0.44} \\
\lscit{}        & \gradient{0.62} & \gradient{0.44} & \gradient{0.88} & \gradient{0.66} & \gradient{0.61} & \gradient{0.44} \\
\midrule
\emph{mistral} & \gradient{0.54} & \gradient{0.46} & \gradient{0.52} & \gradient{0.40} & \gradient{0.68} & \gradient{0.54} \\
\mcits{}       & \gradient{0.54} & \gradient{0.42} & \gradient{0.48} & \gradient{0.34} & \gradient{0.80} & \gradient{0.65} \\
\mcitm{}       & \gradient{0.54} & \gradient{0.41} & \gradient{0.47} & \gradient{0.32} & \gradient{0.81} & \gradient{0.67} \\
\mcit{}        & \gradient{0.44} & \gradient{0.29} & \gradient{0.35} & \gradient{0.20} & \gradient{0.94} & \gradient{0.85} \\
\bottomrule
\end{tabular}

\caption{The AUROC achieved by all the models (rows) at different levels of fine-tuning, from pretrained only to fine-tuned on the full dataset.
    In all settings, the AUROC for models fine-tuned on 3981 and 7861 samples is very close to the results of the fully fine-tuned model. However, the best results are always on the diagonal cells, where the detector and generator models are the same.} %
\label{tab:change_it_fine_tune_detection_scores}
\end{table*}

Finally, for each of the three synthetic datasets generated with \lscit, \lmcit{} and \mcit{}, we compute \emph{log-likelihood} and \emph{DetectGPT} score with
all 12 models: 3 fine-tuned on the full CHANGE-it dataset, 3 fine-tuned on 3981 samples, 3 on 7862 samples, and the original LLMs without any fine-tuning.

\paragraph{Results.} \cref{tab:change_it_fine_tune_detection_scores} suggests that %
the answer to our research question is positive: \textbf{a small subset of fine-tuning samples is indeed sufficient to detect a full fine-tuned model}. A model fine-tuned on only 3\% of the fine-tuning dataset achieves between 86.1\% and 95.4\% of the AUROC measured for the fine-tuned one.

However, this is only effective if the same LLM is both the generator and the detector, see \cref{app:roc_curves} for the ROC curves in this case. Simply fine-tuning different LLMs does not make them similar enough to use one for detecting another. In the case of \mcit{}, it actually gets worse after fine-tuning (we hypothesize that it could be due to differences in tokenization).

While this study focuses on the Italian news, in \cref{app:xsum_proxy_experiments} we perform the same experiments on the XSUM dataset in English, and come to equivalent conclusions: we can detect a fully fine-tuned model with a model that is fine-tuned on a small subset of the whole fine-tuning dataset.

\subsection{Can Ensembling Help?}
\label{sec:llm-id}

As shown in \cref{tab:change_it_fine_tune_detection_scores}, fine-tuning on few samples is sufficient to achieve strong AUROC both with \emph{log-likelihood} and \emph{DetectGPT} on fine-tuned versions of the same LLM -- but detecting fine-tuned versions of different LLMs is harder, and longer fine-tuning does not improve the performance of statistical detection methods%
. This makes the proxy model approach impractical, since in the real world we would not know which base LLM was used.

To address this limitation, a straightforward approach would be ensembling the \emph{DetectGPT} score, which different candidate LLMs assign to a given sample. If texts from a certain source consistently get a high AUROC, this would signal that they are probably synthetic. This is reasonable on the assumption that the CFM is one of the recent high-performing open-source LLMs, and there are not too many of these. Still, to evaluate the potential of this approach we experiment with computing both the mean and max \emph{DetectGPT} score among three LLMs with the same fine-tuning, to understand if this provides a score with higher AUROC.

The results are shown in \cref{tab:multimodel_detectgpt}. %
Neither the average nor the maximum \emph{DetectGPT} score offer a simple solution, but in most cases, one of them yields a significantly higher AUROC than the random guess. We believe that this is overall a promising direction for developing new statistic approaches based on mixing the likelihoods of several models, but, once again, it would be difficult to scale to dozens of candidate LLMs.

\begin{table}[t]
\footnotesize
    \centering
    \begin{tabular}{p{1.4cm}p{1cm}p{1.1cm}p{1.1cm}p{1cm}}%
        \toprule
        CHANGE-it samples & Mode & \lscit{} & \lmcit{} & \mcit{} \\
        \midrule
        \multirow[m]{3}{*}{3981} & max & 0.62 & 0.84 & 0.62 \\
         & mean & 0.66 & 0.73 & 0.68 \\
         & random & 0.63 & 0.63 & 0.64 \\
        \cmidrule{1-5}
        \multirow[m]{3}{*}{7962} & max & 0.58 & 0.83 & 0.68 \\
         & mean & 0.65 & 0.73 & 0.69 \\
         & random & 0.62 & 0.63 & 0.64 \\
        \cmidrule{1-5}
        \multirow[m]{3}{*}{Full} & max & 0.75 & 0.74 & 0.92 \\
         & mean & 0.61 & 0.69 & 0.79 \\
         & random & 0.57 & 0.65 & 0.68 \\
        \bottomrule
    \end{tabular}

    \caption{We experiment with ensembling the \emph{DetectGPT} score measured by models with the same amount of fine-tuning (indicated in the first column). We devise three new scores: \emph{random} is computed by randomly picking the \emph{DetectGPT} score from one of the models, \emph{mean} is computed by taking the average value of all models and \emph{max} by taking the highest value.}%

    \label{tab:multimodel_detectgpt}
\end{table}

\section{Discussion: Are There Practical Solutions to Synthetic Text Detection?}

\label{sec:practicability}

As described in \cref{sec:intro}, the `content farms' for news are already wide-spread, and it is in the public interest to identify such `outlets', as soon as possible. %

One key problem is that we would not know which base LLMs were used, and we would not have access to their token likelihood information. This completely precludes using methods like DetectGPT (\cref{sec:log}). Our proxy model approach (\cref{sec:fine_tuning_detection}) could address the latter problem, but the former is far from being solved (\cref{sec:llm-id}), and will only get harder as more LLMs are published. The offending CFM could also be a `closed' model like GPT-4, to which we would never have white-box access.

Supervised approaches can be used, but, as we showed in \cref{sec:supervised_detection},  they rely on a relatively large\footnote{\citet{verma2023ghostbuster} used 30k samples for English alone.} dataset of texts that were manually identified as CFM texts, and human-written texts to serve as `negative` examples, and it is difficult\footnote{As a binary classification task, the detection of synthetic texts from a particular CFM seems intrinsically very challenging, since ideally we would need the non-synthetic examples to accurately represent all possible non-synthetic sources, and also to be matched with a comparable number of synthetic examples. }
to get the human-written samples right. It would take time, expertise, and resources to develop such a dataset. And we might want to detect a CFM before it publishes even 2K ``news''.
The synthetic examples could also come from multiple CFMs, making the classification task even harder. It is telling that the classifier developed by OpenAI, presumably aiming to detect only OpenAI models, was soon shut down due to low accuracy \cite{KirchnerAhmadEtAl_2023_New_AI_classifier_for_indicating_AI-written_text}.

Concerning watermarking, in the CFM case, it is safe to assume that the deployers of such models would likely try to remove or obfuscate the watermarking, which is relatively easy to do by altering the generation strategy. And the public easy-to-use tools highlighting the statistical `evidence' of the watermark\footnote{E.g. demo by \citet{pmlr-v202-kirchenbauer23a} at \url{https://huggingface.co/spaces/tomg-group-umd/lm-watermarking}} could be used not only by those looking to detect CFMs, but also the CFM operators, to obscure the evidence.

We conclude that at the moment there are no practical options for detecting news `content farms' in the wild, and both the open-source LLM community and providers of `closed' LLMs need to consider ways to address that. We are hopeful that some combination of centralized watermarking effort and further development of detection methods could provide a working solution in the future.

Since the hardest problem is identifying the base LLM, one policy direction to consider would be to (a) mandate watermarking, ideally built-into-model-weights, as a pre-requisite for either commercial deployment of LLMs or their open-source publication, (b) maintain a public watermark detection library, allowing to identify a candidate LLM given a text sample. This would not be bullet-proof, but it would raise both the costs of avoiding detection and the awareness of the problem in the NLP community. %

\section{Conclusion}
\label{sec:discussion}

In this work we have shown how creating a `content farm' generative model for news-like text can be easy, even though we started with a relatively old LLM and a language it was not originally meant for.
After fine-tuning Llama 65B on only 40K Italian news texts, native speakers of Italian have only $\approx 64\%$ accuracy on the synthetic news text detection task. %

We show that the current approaches to automatic text detection, based on token likelihood and supervised classification, outperform human raters in the synthetic text detection, but they would all be impractical in the real world, %
since they require access to the token likelihood information or a large training dataset. We further consider the proxy model approach, and we find that it works well even with little data for fine-tuning, but only if it is known which base LLM was used. Our study highlights the urgency of further work on developing model-agnostic methods of synthetic text detection.

\section{Acknowledgements}
We acknowledge the CINECA award under the ISCRA initiative, for the availability of high-performance computing resources and support IsCb2\_GELATINO (HP10CQRW2J) and IsCb3\_TRAVEL (HP10CY9V7K).

This work was also partially supported by FAIR (PE00000013) project under the NextGenerationEU programme, partially by the PNRR project ITSERR (CUP
B53C22001770006) and partially by the Project PRIN 2022EPTPJ9 (WEMB – “Word EMBeddings: From Cognitive Linguistics to Language Engineering, and Back”), funded by the Italian Ministry of University and Research (MUR). 

The authors’ opinions do not necessarily reflect those of the funding bodies.

\section{Limitations}
\label{sec:limitations}

\paragraph{Generalizability to other languages.} We present a case study on a single language, and do not intend to claim that it is possible to generate plausible-sounding text in any language, by fine-tuning a mostly-English model like Llama. But our results suggest that it \textit{may} be possible, at least for languages with a similar level of coverage in datasets used for training LLMs. More research is needed to establish both the factors impacting the success of such transfer, and better methods to detect synthetic texts.

For the original Llama, according to  \citet[][p.2]{touvron2023llama}, it was exposed to Italian Wikipedia in pre-training. Italian Wikipedia currently has about 500K articles.\footnote{\url{https://en.wikipedia.org/wiki/Wikipedia:Multilingual_statistics}} For other sources included in Llama, such as C4 \cite{raffel_2020_t5}, we cannot exclude the possibility that there was some Italian -- but deliberate effort was made to filter out non-English texts, and so we assume that there was at least not much contamination. Other languages in Llama with about the same amount of Wikipedia data as Italian are Polish and Dutch ($\approx$ 500K articles). In the $\approx$ 400K range there are Spanish and Portuguese, and at about $\approx$ 300K -- Russian and Swedish. A future study could explore how the amount of Wikipedia data, the amount of fine-tuning data, and the typological distance from English impact the success of the transfer.

It is of course also possible and likely that a CFM developer aiming for a specific language would start with an LLM that is multilingual by design, such as BLOOM \cite{ScaoFanEtAl_2022_BLOOM_176BParameter_OpenAccess_Multilingual_Language_Model} or AYA \cite{aryabumi2024aya}, and probably get even better transfer.

\paragraph{Other methods of detecting synthetic text.} In the scope of this paper, we experimented with two statistical detection methods (based on likelihood scores and supervised detection), but there are others, including more sophisticated likelihood based approaches \cite{su-etal-2023-detectllm} that however share the same dependency on the likelihood of the original models, and therefore the core limitations of the methods we experiment with. %
This does not invalidate our conclusions and the general answer to our research question, but it could be expanded in the future work.

\paragraph{Limitations of the human evaluation.}
Our selection of human raters was based solely on Italian as their native language. Future work could investigate whether the results would differ across different occupations and education levels, and with different kinds of synthetic and real articles.

Our human evaluation protocol considers the setting where the model is prompted with the first 30 tokens of a real human-written article, because the model is not trained using the articles headlines but just to generate news, to make the adaptation from English to Italian simpler. Another scenario to be tested in future work is when the model is prompted with headlines (authored by the content farm owner or auto-generated). That could affect the quality of the generated text or the ease of its detection.

Finally, our study focuses on the possibility to create plausible-sounding news-like text that could be used by ``content farms'', rather than text created for specific misinformation campaigns or to spread conspiracy theories. It is possible that, similarly to human-authored fake news, the human raters would be more likely to doubt the authenticity of the article when it had some big factual claims that were easy to check. This factor also  %
remains to be explored in future work. %

The best case scenario with a CFM is that the users would soon realize that the site is fake, and leave -- but even in that case they would already have wasted time and resources, and potentially increased their digital footprint because of tracking on the CFM website. Further possible harms from misinformation, manipulative targeted ads etc. are beyond the scope of this study, and will require a more detailed investigation of various types of factually problematic content, deliberate attempts to introduce certain narratives, awareness and training of the users, etc.

\paragraph{Detection of API-based models.} This study focused on the scenario where the `content farm' used its own model, created by fine-tuning a publicly available high-performing base LLM. It could of course also use an external API service, which would make its task even easier technically.

Famously, GPT-2 \cite{radford2019language} initially came with warnings about it being ``too dangerous to release'', precisely because of the danger of synthetic `news' \cite{Wakefield_2019_Dangerous_AI_offers_to_write_fake_news}. Section 7.4 in the GPT-3 \cite{brown_2020_gpt3} report is dedicated to synthetic news generation and the finding that humans detect such 200-word GPT-3 texts in the 52\%-76\% accuracy range. At that time, too, there was coverage of the dangers of synthetic news \cite[e.g.][]{Mak_2019_When_Is_Technology_Too_Dangerous_to_Release_to_Public,Knight_AI_Can_Write_Disinformation_Now_and_Dupe_Human_Readers}, but no policies ensued.%

Now that OpenAI offers the most popular generative AI services, its proposed solution is its Terms of Service, presumably enforced via constant monitoring of the API use by all users. Its current Terms of Service broadly prohibit ``any harmful, illegal, or abusive activity'', but it is not immediately clear which definitions for `harm' and `legality' must be followed, and whether the news `content farms' websites are covered. The most directly relevant clause currently\footnote{\url{https://openai.com/policies/terms-of-use}, accessed on Feb 10 2023.} seems to be representing ``that Output was human-generated when it was not''. We do not know how this is enforced, and how many bad actors are successfully stopped with API-level controls -- but clearly not all of them.\footnote{For example, Gizmodo identified a fake story about the death of Joe Biden that started with ``\textit{I'm sorry, I cannot complete this prompt as it goes against OpenAI's use case policy on generating misleading content. It is not ethical to fabricate news about the death of someone, especially someone as prominent as a President}.'' \cite{DeGeurin_2023_No_Biden_Isn_Dead_AI_Content_Farms_Are_Pumping_Out_Fake_Stories}}

\section{Broader Impacts}

\paragraph{Impact on society.} This work aims to highlight a potential problem for the information infrastructure of worldwide communities, that may currently consider themselves safe from plausible-looking synthetic text due to the lack of high-quality monolingual models for their languages. We show that a relatively old Llama model, exposed only to Italian Wikipedia and 40K news articles for fine-tuning, is sufficient for generating very plausible-looking synthetic news in Italian, and there are no practical solutions for detecting such text. We hope that this work would spark similar investigations for other languages, and highlight the urgency of development of reliable and model-agnostic methods for detecting synthetic text.

In particular, we hope to draw the attention to the fact that at present, the most authoritative source of information about the extent of the problem with the `content farm' news websites seems to be the aforementioned NewsGuard reports \cite{SadeghiArvanitis_2023_Rise_of_Newsbots_AI-Generated_News_Websites_Proliferating_Online,Tracking_AI-enabled_Misinformation_Over_650_Unreliable_AI-Generated_News_Websites_and_Counting_Plus_Top_False_Narratives_Generated_by_Artificial_Intelligence_Tools}. They are based on extensive expert research and manually vetting different news outlets, and are provided as a paid service. Ideally, the society would be better informed about the scope of the problem\footnote{The fact that number of such websites grew so quickly in the past year (from 49 to 840, see \autoref{sec:intro}) must mean that they are sufficiently profitable. Hence, a large number of people must be mislead \textit{at least} to waste their time and resources on visiting a spammy website, and there are other possible harms (e.g. resulting from misinformation).}, have a reliable public infrastructure for news resources that come from real outlets with editorial responsibility, and taking quick action on the misleading AI-generated websites. %
The problem with the current wave of such `news' originates in the field of NLP, and we hope that our field can also contribute practical solutions to the problems of detecting synthetic texts and assisting with their reporting.

\paragraph{Human and computational resources.} This work is based on the publicly available models \cite{radford2019language, touvron2023llama, touvron2023llama2, raffel_2020_t5, sarti2022it5, jiang2023mistral} and resources \cite{narayan_2018_xsum, changeit_2020}, and documents its emissions (\cref{app:emissions}), annotation procedure and compensation to the human raters (\cref{sec:human}). The code to reproduce our experiments accompanies the submission will be publicly available with the publication of the paper.

\bibliography{custom,lowResLang,anthology_1, anthology_2}

\clearpage

\begin{table*}[t]
\footnotesize
    \centering
\begin{adjustbox}{width=\textwidth}
    \begin{tabular}{cccccc}
        \toprule
        pretrained & \multicolumn{4}{c}{CHANGE-it fine-tuning samples} & HuggingFace model\\
         & 3,981 & 7,862 & 40,000 & 127,392 \\
        \midrule
        \emph{llama-1-7b} & - & - & llama-7b\_it & - & huggyllama/llama-7b\\
        \midrule
        \emph{llama-1-65b} & - & - & llamam-65b\_it & - & huggyllama/llama-65b\\
        \midrule
        \emph{llama-2-7b} & \lscits{} & \lscitm{} & - & \lscit{} & meta-llama/Llama-2-7b\\
        \emph{llama-2-13b} & \lmcits{} & \lmcitm{} & - & \lmcit{} & meta-llama/Llama-2-13b\\
        \emph{mistral} & \mcits{} & \mcitm{} & & \mcit{} & meta-llama/Llama-2-70b\\
        \bottomrule
    \end{tabular}
\end{adjustbox}
    \caption{Model naming based on pretrained model and number of fine-tuning samples. The fine-tuning samples number refers to the training set of CHANGE-it \cite{changeit_2020} dataset. HuggingFace model names correspond to the current links on HuggingFace hub (e.g. for \emph{meta-llama/Llama-2-7b} the pre-trained model comes from \url{https://huggingface.co/meta-llama/Llama-2-7b}).}
    \label{tab:model_naming}
\end{table*}

\appendix

\section{Model Sources and Naming Scheme}
\label{app:model_naming}
All LLMs used in this study are listed in \cref{tab:model_naming}. We also add name of the models in Huggingface-Hub, e.g. for \emph{meta-llama/Llama-2-7b} the pre-trained model comes from \url{https://huggingface.co/meta-llama/Llama-2-7b}.

\section{Fine-tuning Details}

\subsection{Llama 1}
\label{app:finetuningllama1}

We fine-tune both Llama 7b and 65b models on a randomly chosen 40K subset of the CHANGE-it news dataset. The articles are arranged in training sequences composed of 128 tokens, adding subsequent segments one after the other.

The training was performed on 8 nodes, each with 4 V100 GPUs with 16GB VRAM. The effective batch size is 128, real batch size 2, 64 accumulation steps. The maximum learning rate is 0.0005, with a one-cycle scheduler without warmup.

Due to memory constraints, instead of the most effective AdamW optimizer we use simple Stochastic Gradient Descent. We train for 60,000 steps (120,000 samples split into 3 epochs of 40,000) saving checkpoints every 10,000 steps (20,000 samples). This takes approximately 3 days.

\subsection{Llama 2 and Mistral}
\label{app:finetuningllama2}

We fine-tune Llama 2 7b and 13b as well as Mistral models on the full training set of the CHANGE-it news dataset. With a length up to 2048 tokens.

The training is performed on 4 nodes, each with 4 A100 GPUs with 64GB VRAM. The effective batch size is 256, real batch size 4 with 4 accumulation steps. The maximum learning rate is 1.e-5, with a one-cycle scheduler without warm-up.

We use AdamW optimizer with FSDP, we train for three epochs in bfloat16 to limit the memory needed and we perform full fine-tuning without using any parameter efficient technique.

For the fine-tuning on subsets of the CHANGE-it dataset we keep most things equal to the  longer training-set.

For the XSUM dataset the same setting is kept almost identical.

\subsection{Fine Tuning Supervised Synthetic Text Detectors}
\label{app:finetuningclassifier}

\begin{figure}[t]
    \centering
    \includegraphics[width=\linewidth]{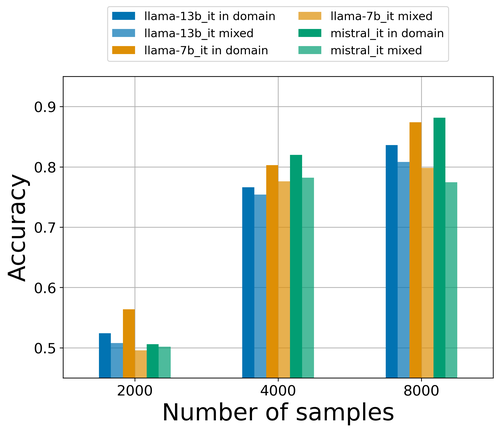}
    \caption{Accuracy of classifier based on RoBERTa-large for human/synthetic text classification task, for synthetic texts generated by three LLMs fine-tuned on CHANGE-it. The classifier was trained on 50\% synthetic texts and either 50\% CHANGE-it texts (\emph{in domain}), or 25\% texts from CHANGE-it and 25\% from DICE (\emph{mixed source}). Classification is only successful at at least 4K labeled samples, and the \emph{mixed source} scenario is consistently more challenging.}
    \label{fig:supervised_detection_roberta}
\end{figure}

We fine-tune two classifiers to identify real or synthetic texts, RoBERTa-large and XLM-RoBERTa-Large. The models are trained on three different dataset size, 2k, 4k, 8k and two data mixing.

The max learning rate is 5.e-5 we use the a batch size of 128 and a linear decaying learning rate without warmup. The rest of the parameters are the daults parameters used by Huggingface Trainer.

The performance is reported in \cref{fig:supervised_detection_xlm} for XLM-RoBERTa and \cref{fig:supervised_detection_roberta} for RoBERTa.

The detailed numerical results are also listed in \cref{tab:supervised_detection}.
\begin{table}[t]
\centering
\footnotesize
\begin{tabular}{lccc}
    \toprule
    generator & n samples  & roberta-large & xlm-large \\
    \midrule
    \multicolumn{4}{c}{In Domain} \\
    \midrule
    \multirow[m]{3}{*}{\lmcit{}} & 2000 & 0.52 & 0.73 \\
     & 4000 & 0.77 & 0.81 \\
     & 8000 & 0.84 & 0.90 \\
    \cmidrule{1-4}
    \multirow[m]{3}{*}{\lscit} & 2000 & 0.56 & 0.59 \\
     & 4000 & 0.80 & 0.86 \\
     & 8000 & 0.87 & 0.92 \\
     \cmidrule{1-4}
     \multirow[m]{3}{*}{\mcit{}} & 2000 & 0.51 & 0.52 \\
     & 4000 & 0.82 & 0.81 \\
     & 8000 & 0.88 & 0.84 \\
     \midrule
     \multicolumn{4}{c}{Mixed Source} \\
     \midrule
     \multirow[m]{3}{*}{\lscit{}} & 2000 & 0.51 & 0.53 \\
    & 4000 & 0.75 & 0.79 \\
    & 8000 & 0.81 & 0.88 \\
    \cmidrule{1-4}
    \multirow[m]{3}{*}{\lmcit{}} &  2000 & 0.50 & 0.61 \\
    &  4000 & 0.78 & 0.86 \\
    &  8000 & 0.80 & 0.90 \\
    \cmidrule{1-4}
    \multirow[m]{3}{*}{\mcit{}} &  2000 & 0.50 & 0.55 \\
    &  4000 & 0.78 & 0.82 \\
    &  8000 & 0.77 & 0.83 \\
    \bottomrule
    \end{tabular}
    \caption{Accuracy achieved by supervised classification models when trained to classify real and synthetic text generated by the various fine-tuned models. We report the accuracy on \emph{in domain} (where all synthetic texts are generated from the same source) and on \emph{mixed source} data (where half of the human written texts come from a different Italian news outlet). Our results suggest that supervised classification of synthetic texts critically depends on the availability of a large (at least 4K in this case) labelled training set, particularly in the \emph{mixed source} scenario.}
    \label{tab:supervised_detection}
\end{table}

\section{Alternative Threshold for Human Evaluation Metric}

\label{app:human_eval_scaled_mean}

To compute the readers accuracy on identifying machine generated texts, we threshold the average score assigned to a sample to obtain a binary label. In \cref{sec:human} we show the results using 3 as a threshold, which is the mean possible rating, but we find that the results are similar when the threshold is the \textbf{scaled mean}: the mean score of all questions in a 100-question survey. \cref{tab:human_eval_scaled_mean} shows all results with this threshold, and they are well aligned with \cref{tab:human_eval_mean}.

\begin{table}[t]
\footnotesize
    \centering
\begin{tabular}{ccc}
\toprule
model & accuracy & std \\
\midrule
\emph{Llama 7B} pretrain & 85.3 & 6.1 \\
\emph{Llama 7B} finetune & 73.7 & 8.8 \\
\midrule
\emph{Llama 65B} pretrain & 72.6 & 3.9 \\
\emph{Llama 65B} finetune & 65.3 & 16.9 \\
\bottomrule
\end{tabular}%
    \caption{Accuracy achieved by human raters in assessing human-written versus machine generated news. We report the overall accuracy and the standard deviation.}
    \label{tab:human_eval_scaled_mean}
\end{table}

\section{Generation examples}
\label{app:examples}

This appendix lists further examples of LLMs pre-trained mostly on English switching from English to Italian when generating the text, as well as the same examples where our CFMs don't do so. We show this for Llama 7b not-pretrained \cref{tab:7B_pretrained_hard}, Llama 7b fine-tuned \cref{tab:7B_finetune_hard}, Llama 65b pre-trained \cref{tab:65B_pretrained_hard} and Llama 65b pre-trained \cref{tab:65B_fine_tune_hard}. The captions also report a break down of the output of the qualitative study we carried out, listing the amount and type of mistakes we spot.

Furthermore, \cref{tab:t5_examples} shows examples of the changes done by it5 when generating the modifications necessary to compute the \emph{DetectGPT} score. These modifications can sometimes lchange meaning in Italian, although when the modified spans are only 1 or 2 words long they keep their original meaning.

\begin{table*}
    \tablefirsthead{%
		\toprule
		   \textbf{Models Switching to English} \\
		\midrule}
	\tablehead{%
		\toprule
		   \textbf{Field} \\
		\midrule}
	\tabletail{%
		\bottomrule
	}
	\tablelasttail{\bottomrule}
\begin{supertabular}{p{14cm}}
        \textbf{\footnotesize Original Text:} {\footnotesize {\color{red} Due sbarchi in sole ventiquattr'ore a Lampedusa. Sono 84 gli immigrati giunti nell'isola tra venerdì e sabato e ospitati all'Hotspot di contrada Imbriacola. Tra loro ci sono anche I tre} unici superstiti dell'ultimo naufragio avvenuto a 50 chilometri a nord est dalla Libia, costato la vita a 117 persone, che erano partite giovedì a bordo di un gommone sgonfiatosi a circa undici ore di navigazione. I passeggeri sono morti annegati, mentre I tre naufraghi sono riusciti a trovare un appiglio fino all'arrivo dei soccorsi italiani. I tre stranieri...} \\
    \midrule
        \textbf{\footnotesize Pre-trained Llama:} {\footnotesize {\color{red} Due sbarchi in sole ventiquattr'ore a Lampedusa. Sono 84 gli immigrati giunti nell'isola tra venerdì e sabato e ospitati all'Hotspot di contrada Imbriacola. Tra loro ci sono anche I tre} naufraghi di Tortuga, recuperati da un mercantile libanese. Non è ancora stata resa nota l'identità nazionale dei naufraghi, ma non mancano inchieste di persone che si sono rivolte ai carabinieri per cercare i loro cari. Lampedusa, 38 migrants - mostly from Tunisia - arrived safely on the island ...} \\
    \midrule
        \textbf{\footnotesize Fine tuned epoch 6 Llama:} {\footnotesize {\color{red} Due sbarchi in sole ventiquattr'ore a Lampedusa. Sono 84 gli immigrati giunti nell'isola tra venerdì e sabato e ospitati all'Hotspot di contrada Imbriacola. Tra loro ci sono anche I tre} bambini rimasti uccisi durante lo stesso sbarcho, due di questi non appartenevano a una famiglia, ma erano stati imbarcati separati dai genitori per il viaggio da Libia verso l'Italia. È stato appunto il maggiore Damiano Maiorino, di stanza a Lampedusa, a scoprire i morti. "Non sono morti cadendo in mare da una barca troppo affollata. Hanno subito commentato ...} \\
    \toprule
        \textbf{\footnotesize Original Text:} {\footnotesize {\color{red} L’ Onu è pronta a fare la sua parte alla conclusione del vertice europeo, ma il segretariato non ha in agenda alcuna azione militare, anche se ci possono essere alcune} similarità con l’operazione in Somalia ". In una intervista a Qn il vicesegretario generale dell’Onu Jan Eliasson apre "all'uso della forza" per fermare gli sbarchi dei clandestini ed evitare nuove stragi nel Mediterraneo. Ma, al tempo stesso, sottolinea "una differenza qualitativa" con l’operazione in Somalia: "L'attività dei pirati era solo criminale, mentre in Libia abbiamo un urgentissimo problema umanitario prima che criminale". Parlando dell’affondamento preventivo dei barconi nei porti libici...} \\
    \midrule
        \textbf{\footnotesize Pre-trained Llama:} {\footnotesize {\color{red} L’ Onu è pronta a fare la sua parte alla conclusione del vertice europeo, ma il segretariato non ha in agenda alcuna azione militare, anche se ci possono essere alcune} iniziative sul campo degli aiuti umanitari", ha affermato un portavoce dell’organizzazione internazionale. Per il segretario generale, "l’ Italia deve avere una posizione di leader ed essere la prima a proporre una risposta alla crisi europea, visto anche che l’ Italia ha avuto un ruolo molto importante nei negoziati sul tema del contingente europeo". UNITED NATIONS, January 31 -- Without waiting for the European Union's decision on...} \\
    \midrule
        \textbf{\footnotesize Fine tuned epoch 6 Llama:} {\footnotesize {\color{red} L’ Onu è pronta a fare la sua parte alla conclusione del vertice europeo, ma il segretariato non ha in agenda alcuna azione militare, anche se ci possono essere alcune} azioni umanitarie" . Così parla il portavoce dell'organizzazione internazionale, Stephane Dujarric, che aggiunge: "Un vertice europeo è una questione interna". La conferma che l'Onu non è pronta a seguito di una intervista rilasciata dal segretario generale aggiunto, Jeffrey Feltman, a radio 24. “Non abbiamo notizie di alcuna iniziativa che possa condurre all'attuazione di un mandato militare contro l'islamic State sul territorio siriano, anche se ritengo che si stia...} \\
    \end{supertabular}

    \caption{Real and generated sentences in Italian from Llama 65B pretrained and Llama 65B fine-tuned on Italian all starting from the same {\color{red} prompt}. The sentences generated by Llama 65B pretrained are not detected by either DetectGPT or log-likelihood, despite being partially written in English -- which could not happen in a real Italian news article.} %
    \label{tab:english_swithch_examples}
\end{table*}

\FloatBarrier
\begin{table*}[b]
    \centering
    \begin{tabular}{p{14cm}}
    \toprule
    \textbf{Llama 7B Pre-Trained} \\
    \toprule
        \textbf{\footnotesize Generated Text:} {\footnotesize {\color{red} I dodici ragazzi rimasti intrappolati con il loro allenatore nella grotta di Tham Luang, nella provincia di Chiang Rai, in Thailandia, per oltre due settimane, hanno lasciato oggi l'ospedale dove}

        erano ricoverati. La polizia ha riferito che i {\color{blue} ragazzi sono stati sottoposti a un test del DNA che ha dimostrato che tutti erano discendenti di due genitori}, ma non vi è nessuna prova che uno o più ragazzi non sia rimasto tra loro durante la fuga dalla grotta. Il...} \\
    \midrule
        \textbf{\footnotesize Translation:} {\footnotesize {\color{red} The 12 boys who were trapped with their coach in the Tham Luang cave in Thailand's Chiang Rai province for more than two weeks left the hospital today where} they were hospitalized. Police reported that {\color{blue} the boys underwent a DNA test that showed that all were descendants of two parents,} but there is no evidence that one or more of the boys did not remain among them while escaping the cave. The} \\
    \midrule
        \textbf{\footnotesize Issue type:} { {\footnotesize \color{blue} Grammatical error.}} \\
    \toprule
        \textbf{\footnotesize Original Text:} {\footnotesize {\color{red} Se dovesse darsi un voto, lei si darebbe un “sette e mezzo”. Sette romani su dieci, però, secondo un sondaggio di Repubblica, non le darebbero nemmeno la sufficienza. Ad un} anno e mezzo dal suo insediamento, è la dimostrazione più esplicita del limpido disaccordo che la popolazione ha con la sua amministrazione. {\color{blue} From the review of A. Kersalme’s “Odysseus” on the site of French journal “La Revue de Belles Lettres” (98th issue, March 2019). From the review of M....}} \\
    \midrule
        \textbf{\footnotesize Translation:} {\footnotesize {\color{red} If she had to give herself a grade, she would give herself a "seven and a half." Seven out of ten Romans, however, would not even give her a sufficiency, according to a Repubblica poll. A} year and a half after taking office, it is the most explicit demonstration of the limpid disagreement the population has with her administration. {\color{blue} From the review of A. Kersalme's "Odysseus" on the site of French journal "La Revue de Belles Lettres" (98th issue, March 2019). From the review of M....}} \\
    \midrule
        \textbf{\footnotesize Issue type:}{ {\footnotesize \color{blue} Factual error.}} \\
    \toprule
        {\footnotesize Generated Text:} {\footnotesize {\color{red} Passerà in carcere da un minimo di 42 anni fino all'ergastolo. Si è chiuso con la lettura di una sentenza durissima da parte del giudice della Superior Court di Los} Angeles. La condanna dura fino al 2019, ma {\color{blue} la rischia se si insulta o rinnega} ancora una volta le sue azioni. Il {\color{blue} procuratore generale della California, Kamala Harris,} ha anche chiesto per lui la revoca di libertà condizionata. Il giovane è stato arrestato a Hong Kong nel maggio 2014....} \\
    \midrule
        {\footnotesize Translation:} {\footnotesize {\color{red} He will spend from a minimum of 42 years to life in prison. It ended with the reading of a very harsh sentence by the Los} Angeles Superior Court judge. The sentence lasts until 2019, but he risks it if he insults or disavows his actions again. {\color{blue} California Attorney General Kamala Harris} also sought revocation of parole for him. The young man was arrested in Hong Kong in May 2014...} \\
    \midrule
        \textbf{\footnotesize Issue type:} { {\footnotesize\color{blue} Grammatically correct, but does not sound natural in Italian and shows factual errors.}} \\
    \bottomrule
    \end{tabular}
    \caption{Examples of mistakes made by pretrained Llama 7B. The smaller Llama 7b without fine-tuning shows several kinds of weaknesses generating both factual to grammatical errors. Out of 25 samples manually examined for this model, we find 0 contradicting the prompt, 5 switching to English, 8 unnatural Italian expressions and 6 grammatical errors.}
    \label{tab:7B_pretrained_hard}
\end{table*}

\FloatBarrier
\begin{table*}[b]
    \centering
    \begin{tabular}{p{14cm}}
    \toprule
    \textbf{Llama 7B Fine-Tuned} \\
    \toprule
        \textbf{\footnotesize Generated Text:} {\footnotesize {\color{red} Passa per sette punti il programma di Matteo Salvini per bloccare I flussi e dare maggiore sicurezza al Paese. Il blocco dei porti a navi che battono bandiera di uno} stato membro dell'Unione europea e non il Regno Unito, come annunciato da Salvini, sarebbe contro il diritto e il trattato, {
        \color{blue} secondo l'ex ministro della Giustizia Franco Frattini}. "Vi è un'interpretazione di diritto internazionale. A me pare che sia totalmente sbagliata", ha detto il politico liberale di Palazzo Chigi. "Sotto...} \\
    \midrule
        \textbf{\footnotesize Translation:} {\footnotesize {\color{red} It passes for seven points in Matteo Salvini's program to stop the flows and give more security to the country. Blocking ports to ships flying the flag of a} European Union member state and not the United Kingdom, as announced by Salvini, would be against law and treaty, {\color{blue} according to former Justice Minister Franco Frattini}. "There is an interpretation of international law. It seems to me that it is totally wrong," said the liberal politician from Palazzo Chigi. "Under...} \\
    \midrule
        \textbf{\footnotesize Issue type:} {\footnotesize {\color{blue} Factual errors.}} \\
    \toprule
        \textbf{\footnotesize Original Text:} {\footnotesize {\color{red} Alla vigilia dell’incontro, a Bruxelles, tra Jean-Claude Juncker e il presidente del Consiglio Giuseppe Conte, che sarà accompagnato dal ministro dell’Economia Tria, il commissario Ue Pierre Moscovici usa tona concilianti} per suggerire che l’Italia non dovrebbe fare la spola tra l’Europa e le banche a credito. "Sarà difficile", dice il francese, "se la Banca centrale germanica non pone il suo veto {\color{blue} alla pesca in una sola mano per} un paese come l’Italia che sta in cerca di finanziamenti". "La...} \\
    \midrule
        \textbf{\footnotesize Translation:} {\footnotesize {\color{red} On the eve of a meeting in Brussels between Jean-Claude Juncker and Prime Minister Giuseppe Conte, who will be accompanied by Economy Minister Tria, EU Commissioner Pierre Moscovici uses conciliatory tones} to suggest that Italy should not shuttle between Europe and banks on credit. "It will be difficult," says the Frenchman, "if the German Central Bank does not {\color{blue} veto single-handed fishing for a country} like Italy that is looking for financing." "The...} \\
    \midrule
        \textbf{\footnotesize Issue type: }{\footnotesize {\color{blue} Unnatural expression.}} \\
    \toprule
        {\footnotesize Generated Text:} {\footnotesize {\color{red} Forti temporali si sono abbattuti nella mattinata di oggi su Milano e sul nord-ovest della Lombardia, spazzando via il caldo degli ultimi giorni. Il nubifragio ha creato qualche rallentamento sulle} strade. Poche colture e alcune piantine. Questi sono i principali danni dei fenomeni atmosferici che hanno colpito l’Italia. L’Italia centro-meridionale ha visto {\color{blue} 304.363 colture} danneggiate, quasi {\color{blue} 15mila piantine} e {\color{blue} 1.532 migliaia di alberi}, una pioggia netta di oltre 10mila mila chili sul versante del Po e una coltura danneggiata...} \\
    \midrule
        {\footnotesize Translation:} {\footnotesize {\color{red} Strong thunderstorms hit Milan and northwestern Lombardy in the morning today, sweeping away the heat of the past few days. The cloudburst created some slowdowns on} roads. Few crops and a few seedlings. These are the main damages of the weather phenomena that hit Italy. South-central Italy saw {\color{blue} 304,363 crops damaged}, nearly {\color{blue} 15,000 seedlings} and {\color{blue} 1,532,000 trees}, a net rainfall of more than 10,000,000 kilograms on the Po River side, and a damaged crop...} \\
    \midrule
        {\footnotesize Issue type: } {\footnotesize{\color{blue} Factual error.}} \\
    \bottomrule
    \end{tabular}
    \caption{Examples of mistakes made by Llama 7B fine-tuned on Italian. The smaller Llama 7b after fine-tuning generates flowing text but the facts are anyway less accurate than for larger models. Out of 25 samples manually examined for this model, we find 5 contradicting the prompt, 0 switching to English, 5 unnatural Italian expressions and 6 grammatical errors.}
    \label{tab:7B_finetune_hard}
\end{table*}

\FloatBarrier
\begin{table*}[b]
    \centering
    \begin{tabular}{p{14cm}}
    \toprule
        \textbf{Llama 65B Pre-Trained} \\
    \toprule
        \textbf{\footnotesize Generated Text: }{\footnotesize {\color{red} In centro Italia continuano le scosse e il numero dei morti sale. Paesi afoni e distrutti, palazzi accartocciati, mozziconi di chiese, rovine sinistre che fissano le piazze. E I corpi} stanno fuori {\color{blue} dalle case, senza casa, senza neanche la tomba che li ricuopre}. Le indagini del Giornalista Enrico Lucci e le testimonianze dei superstiti. The BBC's Gavin Lee reports from Norcia, the central Italian town hardest hit by the quake. Italian Prime Minister Matteo Renzi has visited one of...} \\
    \midrule
        \textbf{\footnotesize Translation: }{ \footnotesize { \color{red} In central Italy, the tremors continue and the death toll rises. Aphonious and destroyed villages, crumpled buildings, church butts, sinister ruins staring into squares. And The bodies} stand {\color{blue} outside homes, homeless, without even the grave to cover them}. Journalist Enrico Lucci's investigation and survivor testimony. The BBC's Gavin Lee reports from Norcia, the central Italian town hardest hit by the quake. Italian Prime Minister Matteo Renzi has visited one of...} \\
    \midrule
        \textbf{\footnotesize Issue type: }{\footnotesize {\color{blue} Grammatically correct, but does not sound natural in Italian.}} \\
    \toprule
        \textbf{\footnotesize Original Text:} {\footnotesize {\color{red} Due sbarchi in sole ventiquattr'ore a Lampedusa. Sono 84 gli immigrati giunti nell'isola tra venerdì e sabato e ospitati all'Hotspot di contrada Imbriacola. Tra loro ci sono anche I tre} naufraghi di Tortuga, recuperati da un mercantile libanese. Non è ancora stata resa nota l'identità nazionale dei naufraghi, ma non mancano inchieste di persone che si sono rivolte ai carabinieri per cercare I loro cari. Lampedusa, {\color{blue} 38 migrants - mostly from Tunisia - arrived safely on the island in...}} \\
    \midrule
        \textbf{\footnotesize Translation:} {\footnotesize {\color{red} Two landings in just twenty-four hours on Lampedusa. A total of 84 immigrants arrived on the island between Friday and Saturday and were housed at the contrada Imbriacola Hotspot. Among them are also The three} shipwrecked Tortuga, recovered by a Lebanese merchant ship. The national identity of the castaways has not yet been released, but there is no shortage of inquiries from people who have turned to the Carabinieri to search for their loved ones. Lampedusa, {\color{blue} 38 migrants - mostly from Tunisia - arrived safely on the island in...}} \\
    \midrule
        \textbf{\footnotesize Issue type: }{\footnotesize {\color{blue} Switch to English}}\\
    \toprule
        \textbf{\footnotesize Generated Text: }{\footnotesize {\color{red} I numeri fanno spavento. Cinquemila disperati sbarcati solo nelle ultime quarantott'ore, quasi 65mila nei primi sei mesi dell'anno. Che l' operazione "Mare Nostrum", varata dopo la strage di Lampedusa, fosse} {\color{blue} riuscita a fermare il flusso di barche di migranti}, lo abbiamo rilevato già da pochi mesi. Che la spinta verso l'Italia del prossimo potesse crescere di conseguenza, lo sapevamo. Che si dovesse progettare una strategia efficace e responsabile, in grado di dire dove e come, giustamente, stava l' opportunità...} \\
    \midrule
        \textbf{\footnotesize Translation:}{\footnotesize {\color{red} The numbers are frightening. Five thousand desperate people landed in the last forty-eight hours alone, nearly 65,000 in the first six months of the year. That operation "Mare Nostrum," launched after the Lampedusa massacre, had} {\color{blue} succeeded in stopping the flow of migrant boats}, we noted a few months ago. That the push to Italy of the next could grow as a result, we knew. That an effective and responsible strategy had to be designed that could tell where and how, rightly, the 'opportunity lay....} \\
    \midrule
        \textbf{\footnotesize Issue type: }{\footnotesize {\color{blue} Contradicts the prompt.}} \\
    \bottomrule
    \end{tabular}
    \caption{Examples of subtle mistakes made by pretrained Llama 65B. While the quality of the language is almost on par with the prompts, written by professional journalists (for these short examples), the content is largely made up although in a realistic way. Out of 25 samples manually examined for this model, we find 2 contradicting the prompt, 3 switching to English, 7 unnatural Italian expressions and 2 grammatical errors.}
    \label{tab:65B_pretrained_hard}
\end{table*}

\FloatBarrier
\begin{table*}[b]
    \centering
    \begin{tabular}{p{14cm}}
    \toprule
        \textbf{Llama 65B Fine-Tuned} \\
    \toprule
        \textbf{\footnotesize Generated Text:} {\footnotesize {\color{red} Gabriele Gravina è stato eletto presidente della Figc. Ha ottenuto il 92,2\% dei voti (al primo turno serviva il 75\%). Si chiude la gestione commissariale di Roberto Fabbricini, durata 9} mesi. L'ultimo presidente eletto era Carlo {\color{blue} Tavecchio nel 2014, che aveva ottenuto il 63\%. Gravina: "Riconoscendomi orgogliosamente l'apostrofe di figlio della Sicilia e della Normandia che le mie origini e la mia vita mi fanno, ho iniziato la mia esperienza nel calcio. Un calcio che ho giocato da bambino,...}} \\
    \midrule
        \textbf{\footnotesize Translation:} {\footnotesize {\color{red} Gabriele Gravina has been elected president of Figc. He has obtained 92.2\% of the votes (on first round he needed 76\%), This ends the Roberto Fabbricini's technical administration, which lasted 9} months. The last president to be elected was Carlo {\color{blue} Tavecchio in 2014, he scored 63\%. Gravina: "Proudly acknowledging the apostrophe of son of Sicily and Normandy that my origins and my life make me, I began my experience in soccer. A soccer I played as a child,..."}} \\
    \midrule
        \textbf{\footnotesize Issue type: } {\footnotesize  {\color{blue} Factual errors.}} \\
    \toprule
        \textbf{\footnotesize Original Text:} {\footnotesize {\color{red} “È superficiale dire che è risorta la Dc ”. Parola di Francesco Rutelli che, intervistato dal Corriere, ripercorre le tappe della vita della Margherita, partito in cui hanno militato sia} lui che Bertinotti. Rutelli l’ha lasciato da qualche anno e adesso è presidente dei {\color{blue} Liberali per l’Italia,} la lista che recentemente ha lanciato al Senato. E Bertinotti è tornato alla Dc per partecipare alle primarie del partito di D’Alema e Bersani, l’unico che ha il voto dei sostenitori della...} \\
    \midrule
        \textbf{\footnotesize Translation:} {\footnotesize {\color{red} “It is superficial to say that the DC has risen again." Word of Francesco Rutelli, who, interviewed by Corriere, traces the stages of the life of Margherita, a party in which both} he and Bertinotti militated. Rutelli left it a few years ago and is now president of { \color{blue} Liberals for Italy,} the list he recently launched in the Senate. And Bertinotti returned to the DC to participate in the primaries of D'Alema and Bersani's party, the only one that has the supporter vote of the...} \\
    \midrule
        \textbf{\footnotesize Issue type: }{\footnotesize {\color{blue} Factual errors.}}\\
    \toprule
        \textbf{\footnotesize Generated Text:} {\footnotesize {\color{red} Dopo 19 giorni, grazie all'accordo trovato dall'Europa, I 49 migranti di Sea Watch e Sea Eye sono sbarcati nel porto maltese di La Valetta, dove sono stati trasportati a bordo} di autobus. Il {\color{blue} ministro Migranti e Strade del Popolo} Gabriele Toccafondi nei giorni scorsi ha incontrato a Parigi il ministro degli Esteri francese Jean-Yves Le Drian, il ministro dell'interno Christophe Castaner, il ministro della giustizia Nicole Belloubet e il ministro dell'istruzione Jean-Michel Blanquer. Si tratta di una missione di...} \\
    \midrule
        \textbf{\footnotesize Translation:} {\footnotesize {\color{red} After 19 days, thanks to the agreement found by Europe, The 49 migrants from Sea Watch and Sea Eye landed in the Maltese port of La Valetta, where they were transported aboard} buses. {\color{blue} Migrants and People's Roads Minister} Gabriele Toccafondi in recent days met in Paris with French Foreign Minister Jean-Yves Le Drian, Interior Minister Christophe Castaner, Justice Minister Nicole Belloubet, and Education Minister Jean-Michel Blanquer. This is a mission of...} \\
    \midrule
        \textbf{\footnotesize Issue type: }{\footnotesize {\color{blue}  Factual errors.}} \\
    \bottomrule
    \end{tabular}
    \caption{Examples of subtle mistakes made by Llama 65B fine-tuned on Italian. While the quality of the language is almost on par with the prompts, written by professional journalists (for these short examples), the content is largely made up although in a realistic way. Out of 25 samples manually examined for this model, we find 0 contradicting the prompt, 0 switching to English, 1 unnatural Italian expressions and 4 grammatical errors.}
    \label{tab:65B_fine_tune_hard}
\end{table*}

\FloatBarrier
\subsection{Modifications from IT5}

The modification from IT5 to compute the DetectGPT score don't necessarily need to have a meaning as they are only used to normalize the log-likelihood values, however, it is interesting to note that often they don't disrupt the sentence meaning, and that they can create new sentences, although with similar meaning, that can themselves be useful as synthetic data. We release 600k of these sentences that are half modifications of the original news and half modifications of the synthetic texts generated by our fine-tuned models.

\FloatBarrier
\begin{table*}
\footnotesize
    \tablefirsthead{%
		\toprule
		   \textbf{Field} & \textbf{Content} \\
		\midrule}
	\tablehead{%
		\toprule
		   \textbf{Field} & \textbf{Content} \\
		\midrule}
	\tabletail{%
		\bottomrule
	}
	\tablelasttail{\bottomrule}
\begin{center}
    \begin{supertabular}{lp{12cm}}
    Original Text: & Dato il seguente titolo di un articolo di giornale scrivi l'articolo. \#\#\# Titolo:"Nel Pd niente conte interne". Ma su un sito i nomi dei pro M5S \#\#\# Articolo: Alla vigilia della direzione che potrebbe, ma non necessariamente, sancire la separazione con gli scissionisti guidati da Renzi, il Pd cerca di chiudere la partita che tra poco si apre. La maggioranza del partito, insistono al Nazareno, per ora "non intende partecipare ad un conto interne", \\
\midrule
    Translation & Given the following title of a newspaper article, write the article. \#\#\# Title:"No internal counts in the Democratic Party". But on a site the names of the M5S pros \#\#\# Article: On the eve of the direction that could, but not necessarily, sanction the separation with the splinterists led by Renzi, the Democratic Party tries to close the game that will soon open . The majority of the party, the Nazareno insists, for now "does not intend to participate in an internal account"\\
\midrule
    Modification 1: & Dato il seguente titolo di {\color{red} anteprima} di giornale scrivi l\'articolo.\#\#\# Titolo: "Nel Pd niente {\color{red} esodati}, Ma su un sito I nomi dei pro M5S \#\#\# Articolo: Alla vigilia della direzione che potrebbe, ma non necessariamente, sancire la {\color{red} rottura definitiva} con gli scissionisti guidati da Renzi, il pd cerca {\color{red} il suo partner di governo dietro la partita m5s che da} poco si apre. La maggioranza del partito, insistono al Nazareno, per ora "non intende partecipare ad un conto interne"\\
\midrule
    Translation: & Given the following newspaper {\color{red} preview} title, write an article.\#\#\# Title: "No {\color{red} exoduses} in the Democratic Party, but on a site the names of the M5S pros \#\#\# Article: On the eve of the direction that could , but not necessarily, sanction {\color{red} the definitive break} with the splinterists led by Renzi, the Democratic Party is {\color{red} looking for its government partner behind the m5s match that has recently opened}. The majority of the party, the Nazareno insists, for now "does not intend to participate in an internal account" \\
\midrule
    Modification 2: & Dato il seguente titolo di un articolo di giornale scrivi l\'articolo. \#\#\# Titolo: "Nel Pd niente conte interne". Ma su un {\color{red} annuncio già} I nomi dei pro M5S \#\#\# Articolo: Alla vigilia {\color{red} al voto} che potrebbe, ma non necessariamente, sancire la separazione con {\color{red} I dem} guidati da Renzi, il pd cerca di chiudere la partita che tra poco si apre. La maggioranza del partito, insistono al Nazareno, {\color{red} è ferma sulla linea del "non cedere} ad un conto interne"\\
\midrule
    Translation: & Given the following title of a newspaper article, write the article. \#\#\# Title:"No internal counts in the Democratic Party". But {\color{red} on an announcement already} the names of the M5S pros \#\#\# Article: On the eve of the {\color{red} vote} that could, but not necessarily, sanction the separation with {\color{red} the democrats}, the Democratic Party tries to close the game that will soon open . The majority of the party, the Nazareno insists, {\color{red} is fixed on the idea of "not giving in to} an internal account"\\
    \end{supertabular}
    \caption{Examples of Italian text modifications from IT5 models, used to compute the DetectGPT score. We show the translation and two modifications, in {\color{red} red the chunks that have been replaced}. Depending on the length the modifications can alter the meaning of the original sentences, however they work to normalize a sentence likelihood and compute \emph{DetectGPT}.}
    \label{tab:t5_examples}
    \end{center}
\end{table*}

\FloatBarrier
\clearpage

\begin{figure}[t]
    \centering
    \begin{subfigure}{0.48\linewidth}
        \includegraphics[width=\linewidth]{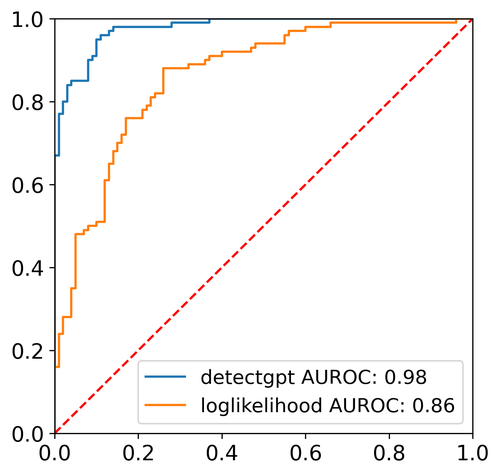}
        \subcaption{GPT2}
        \label{subfig:roc_gpt2}
    \end{subfigure}
    \begin{subfigure}{0.48\linewidth}
        \includegraphics[width=\linewidth]{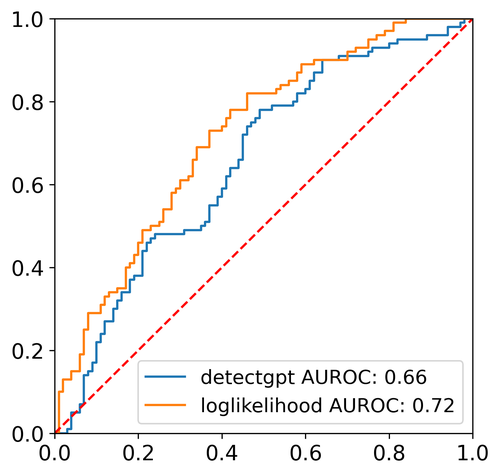}
        \subcaption{Llama 13B}
        \label{subfig:roc_llama13b}
    \end{subfigure}
    \caption{ROC curve for DetectGPT and log-likelihood. In (\subref{subfig:roc_gpt2}) for the GPT2 model over 100 samples from xsum (coherent with \cite{mitchell_detectgpt_2023}), in (\subref{subfig:roc_llama13b}) for llama13b model over 100 samples from xsum.}
    \label{fig:auroc}
\end{figure}

\section{Generalizability to English}
\label{app:xsum_log_experiments}

\subsection{Synthetic Text Detection Based on Token Likelihoods}

\begin{table}[t]
    \centering
    \adjustbox{max width=\linewidth}{%
    \begin{tabular}{ccccc}
    \toprule
        & \multicolumn{3}{c}{\emph{DetectGPT}} & \emph{log-likelihood} \\
        & t5-base & t5-3b & t5-11b &  \\
    \midrule
        Llama 13b & 66\% & 71\% & 75\% & 78\% \\
        Llama 65b & - & 62\% & 66\% & 70\% \\
    \bottomrule
    \end{tabular}}
    \caption{AUROC achieved by \emph{DetectGPT} (varying bootstrapping model) and \emph{log-likelihood} on the xsum data-set for Llama 13B and Llama 65B.}
    \label{tab:auroc_eng}
\end{table}

\begin{table}[t]
\footnotesize
    \centering
    \adjustbox{max width=\linewidth}{%
    \begin{tabular}{ccc}
    \toprule
    Temperature & \emph{DetectGPT} & \emph{log-likelihood} \\
    \midrule
    0.6 & 48\% & 86\% \\
    0.8 & 63\% & 73\% \\
    1.0 & 77\% & 52\% \\
    \bottomrule
    \end{tabular}}
    \caption{AUROC achieved by \emph{DetectGPT} and \emph{log-likelihood} on Llama 13B varying the temperature used while generating the synthetic sentences.}
    \label{tab:auroc_temperature}
\end{table}

\begin{table}[t]
\footnotesize
\centering
\begin{adjustbox}{width=\linewidth}
\begin{tabular}{p{2.3cm}cccc}
\toprule
generator & \multicolumn{2}{c}{llama-2-7b\_xsum} & \multicolumn{2}{l}{mistral\_xsum} \\
 & dGPT & llh & dGPT & llh \\
\midrule
llama-2-7b & \gradient{0.82} & \gradient{0.63} & \gradient{0.65} & \gradient{0.52} \\[2ex]
llama-2-7b\_xsum (995 samples) & \gradient{0.86} & \gradient{0.69} & \gradient{0.65} & \gradient{0.53} \\
llama-2-7b\_xsum & \gradient{0.89} & \gradient{0.75} & \gradient{0.65} & \gradient{0.54} \\ [2ex]
\midrule
Mistral-7B-v0.1 & \gradient{0.59} & \gradient{0.41} & \gradient{0.77} & \gradient{0.59} \\ [2ex]
mistral\_xsum (995 samples) & \gradient{0.57} & \gradient{0.44} & \gradient{0.85} & \gradient{0.75} \\ [2ex]
mistral\_xsum & \gradient{0.53} & \gradient{0.43} & \gradient{0.95} & \gradient{0.92} \\ [2ex]
\bottomrule
\end{tabular}
\end{adjustbox}
\caption{The AUROC achieved by all the models (rows) at different levels of fine-tuning, from pretrained only to fine-tuned on the full dataset. In \textbf{bold} the higher AUROC in each column.}
\label{tab:xsum_fine_tune_detection_scores}
\end{table}

To confirm that our setup in \autoref{sec:log} is comparable to the previous efforts, we start by replicating a core result by \citet{mitchell_detectgpt_2023}: on xsum data \citep{narayan_2018_xsum}, and using different versions of \emph{t5} \citep{raffel_2020_t5}
DetectGPT outperforms the \emph{log-likelihood} in detecting GPT2 text \cite{radford2019language} (see \cref{subfig:roc_gpt2}).

We apply the same methodology to sentences obtained using Llama 13B and Llama 65B. For Llama, we were unable to get \emph{DetectGPT} to achieve a higher AUROC than the \emph{log-likelihood}. We believe this to be due to the stronger performance of Llama compared to the t5 model used to generate new sentences (\cref{subfig:roc_llama13b}).
This suggests that English text generated by Llama is harder to detect.

To test this hypothesis we measure the importance of the bootstrap model in this case. \cref{tab:auroc_eng} shows the AUROC of \emph{DetectGPT} depending on the bootstrap model, a larger t5 model leads to higher AUROC.

We repeat the experiment with both DetectGPT and log-likelihood at various temperature settings (0.6, 0.8, 1.0), and we find a strong sensitivity to this hyper-parameter, which merits further investigation.

To establish a fair comparison with DetectGPT while testing Llama, we perform an ablation study based on varying the temperature used in generation.

\Cref{tab:auroc_temperature} shows different AUROC values for different temperatures. It appears that there is a strong sensitivity of the detection methodologies to this hyper-parameter, which merits further investigation in future work. The value 0.8 where \emph{DetectGPT} and \emph{log-likelihood} are more aligned, is also the value reported in the Llama repository.

\subsection{Detecting CFMs with Proxy Models}
\label{app:xsum_proxy_experiments}

We repeat the experiments done for the CHANGE-it dataset also for the XSUM dataset \cite{narayan_2018_xsum}.

That is we fine-tuned \emph{llama-2-7b} and \emph{mistral} on the full XSUM training set. We generate 2 datasets with 1k synthetic texts generated with
each of the fine-tuned models and 1k samples from the xsum test set (less than for CHANGE-it to limit compute costs). Then we also fine-tune these two models
on a small subset of the trainig set, 995 samples and finally we compute \emph{DetectGPT} achieved by these 6 models,
\emph{llama-2-7b} and \emph{mistral} pre-trained, \emph{llama-2-7b\_xsum\_995} and \emph{mistral\_xsum\_995} fine-tuned on 995 samples and
\emph{llama-2-7b\_xsum} and \emph{mistral\_xsum} fine-tuned on the xsum dataset.

\Cref{tab:xsum_fine_tune_detection_scores} shows the results that closely match those for CHANGE-it shown in \cref{tab:change_it_fine_tune_detection_scores}, namely that
fine-tuning also on a small set of the same domain leads to high AUROC while if models come from different pre-trained ones the performance is low.

\FloatBarrier

\section{Prompting details}
\label{app:prompting}
\autoref{fig:prompt} shows the exact prompt that was used for generating the synthetic texts in \autoref{sec:fine_tuning_detection}. In the prompt, we retain a few initial tokens of the original article, ensuring that the prompt never exceeded 30 words in total.

\begin{figure}[ht]
\begin{mdframed}[backgroundcolor=mygray,linewidth=0pt]
\footnotesize
"""Given the following article title, generate the article.\\
\#\#\# Title:\\
{\color{gray}\{title\}}\\
\#\#\# Article:\\
{\color{gray}\{article\}}"""
\end{mdframed}
\caption{Prompt used for generating news-like texts in \autoref{sec:fine_tuning_detection}.}
\label{fig:prompt}
\end{figure}

\section{Receiver Operating Characteristic (ROC) Curves}
\label{app:roc_curves}

The receiver operating curve is a measure to understand how setting a threshold over a score would influence the true positive rate (TPR, the numer of instances marked as positive divided by the total number of positives)
and the false positive rate (FPR, the number of false positive divided by the number of all negatives). That means how setting all
instances with a score above a certain threshold as positive would influence the number of true positive and false positive.

Given that when we set as a threshold the maximum value of a score all instances are classified as negative TPR equals 0 and FPR equals 0,
viceversa when it is set to the minimum TPR equals 1 and FPR equals 1. The ROC curve is a plot of the TPR against the FPR for different thresholds.

\begin{figure*}
    \centering
    \begin{subfigure}[b]{0.20\textwidth}
        {\Large Llama-2-13b\_it}
        \includegraphics[width=\textwidth]{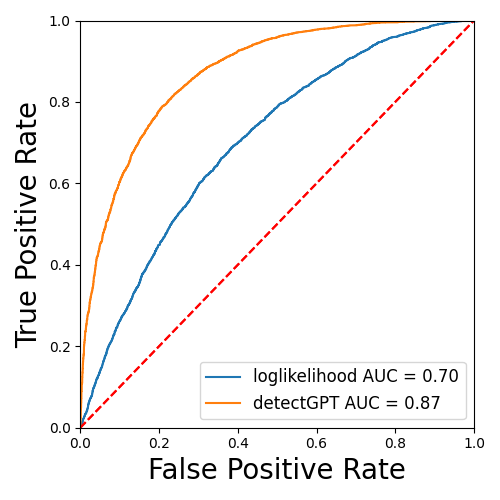}
        \caption{\lmcit{}}
        \label{fig:subfig1}
    \end{subfigure}
    \begin{subfigure}[b]{0.20\textwidth}
        \includegraphics[width=\textwidth]{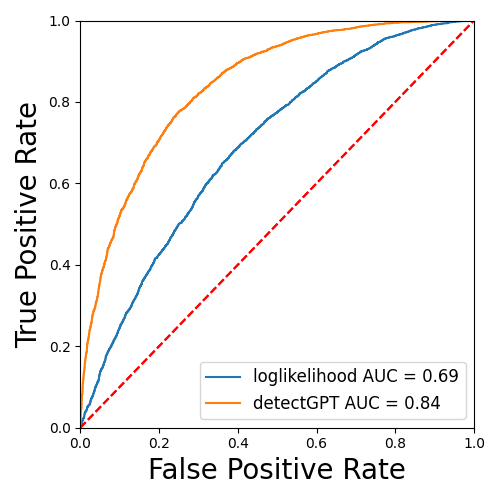}
        \caption{\lmcits{}}
        \label{fig:subfig2}
    \end{subfigure}
    \begin{subfigure}[b]{0.20\textwidth}
        \includegraphics[width=\textwidth]{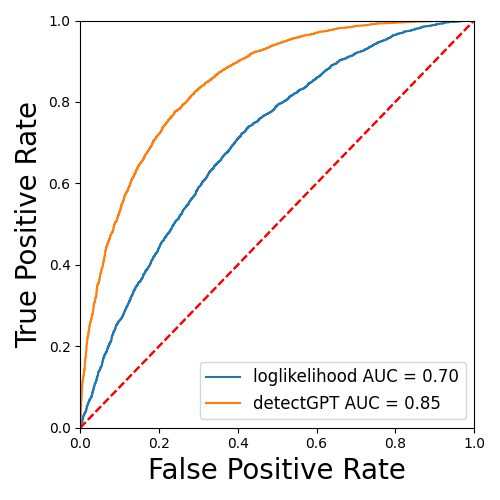}
        \caption{\lmcitm{}}
        \label{fig:subfig3}
    \end{subfigure}
    \begin{subfigure}[b]{0.20\textwidth}
        \includegraphics[width=\textwidth]{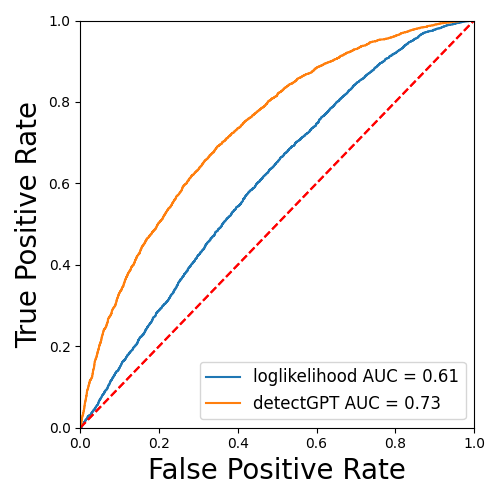}
        \caption{\emph{llama-2-13b}}
        \label{fig:subfig4}
    \end{subfigure}

\vspace{0.5cm}
    \begin{subfigure}[b]{0.20\textwidth}
        {\Large Llama-2-7b\_it}
        \includegraphics[width=\textwidth]{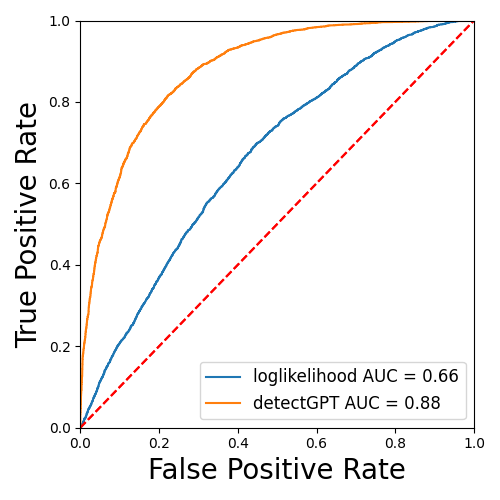}
        \caption{\lscit{}}
        \label{fig:subfig5}
    \end{subfigure}
    \begin{subfigure}[b]{0.20\textwidth}
        \includegraphics[width=\textwidth]{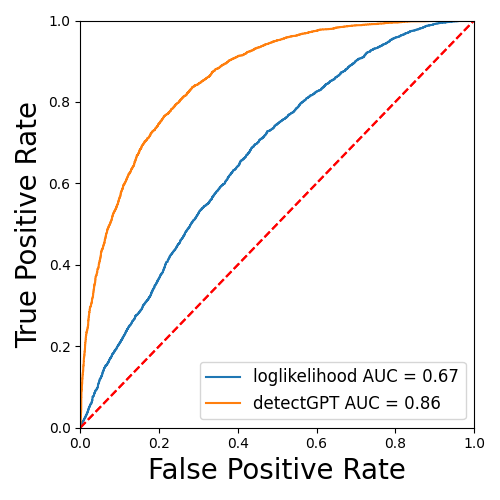}
        \caption{\lscits{}}
        \label{fig:subfig6}
    \end{subfigure}
    \begin{subfigure}[b]{0.20\textwidth}
        \includegraphics[width=\textwidth]{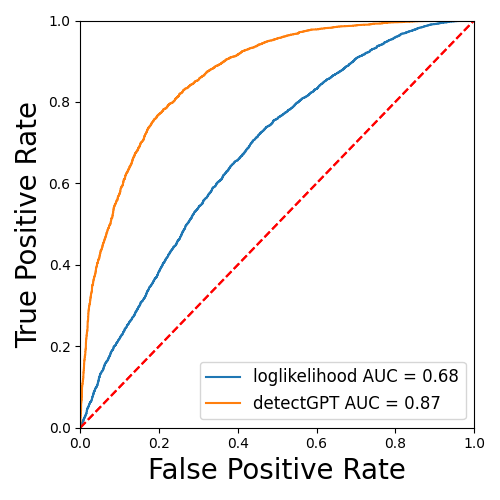}
        \caption{\lscitm{}}
        \label{fig:subfig7}
    \end{subfigure}
    \begin{subfigure}[b]{0.20\textwidth}
        \includegraphics[width=\textwidth]{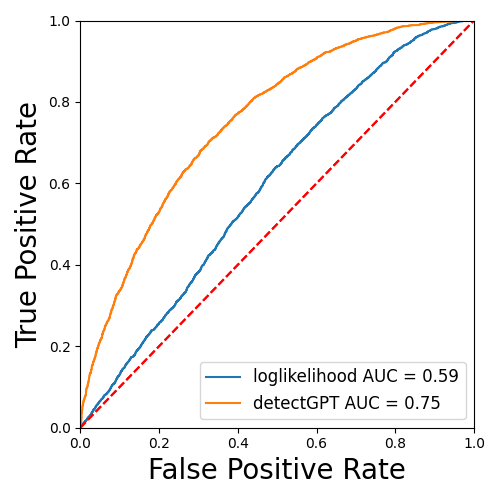}
        \caption{\emph{llama-2-7b}}
        \label{fig:subfig8}
    \end{subfigure}

\vspace{0.5cm}
    \begin{subfigure}[b]{0.20\textwidth}
        {\Large Mistral\_it}
        \includegraphics[width=\textwidth]{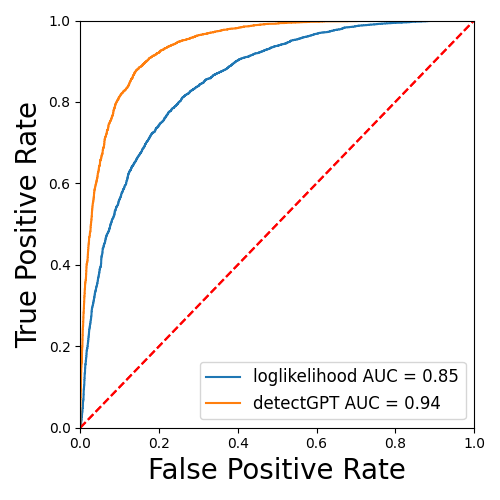}
        \caption{\lmcit{}}
        \label{fig:subfig9}
    \end{subfigure}
    \begin{subfigure}[b]{0.20\textwidth}
        \includegraphics[width=\textwidth]{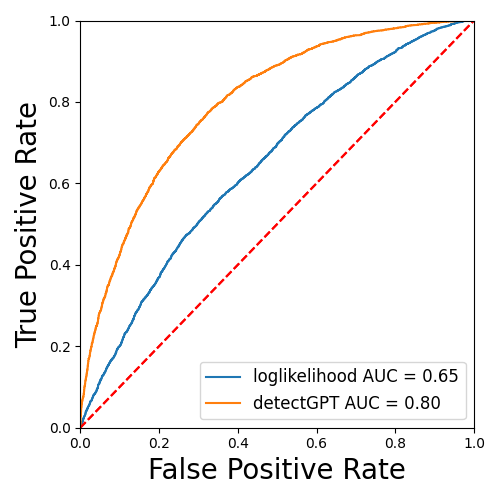}
        \caption{\lmcits{}}
        \label{fig:subfig10}
    \end{subfigure}
    \begin{subfigure}[b]{0.20\textwidth}
        \includegraphics[width=\textwidth]{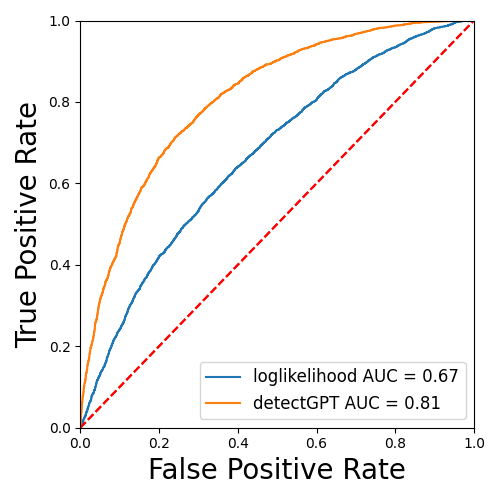}
        \caption{\lmcitm{}}
        \label{fig:subfig11}
    \end{subfigure}
    \begin{subfigure}[b]{0.20\textwidth}
        \includegraphics[width=\textwidth]{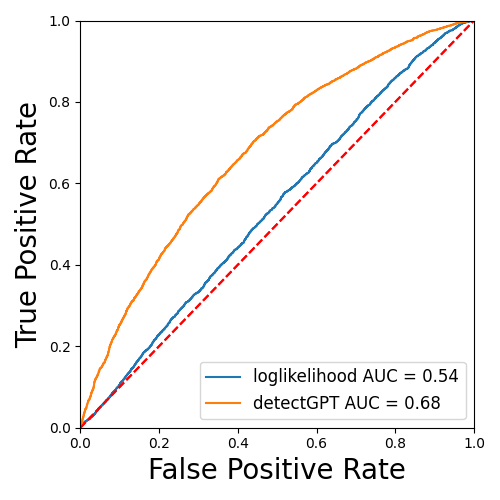}
        \caption{\emph{mistral}}
        \label{fig:subfig12}
    \end{subfigure}
    \caption{The ROC curves for the proxy models when evaluated on data generated by models fine-tuned starting from the same pre-trained model.}
    \label{fig:proxy_model_auroc}
\end{figure*}

To show further details about the proxy models AUROC values shown in \cref{tab:change_it_fine_tune_detection_scores}.
\Cref{fig:proxy_model_auroc} shows the ROC curves for the case when the proxy models come from the same pre-trained one.

We can see tha computing the AUROC over 10k examples leads to a smooth curve and that the shape is very consistent across different amounts of fine-tuning
futher validating the strngth of proxy models.

\FloatBarrier

\clearpage
\FloatBarrier

\section{Computational Costs}
\label{app:emissions}
The fine-tuning run of our `CFM' Llama (\cref{sec:llama}) lasted 5 days, as we wasted approximately 2 days due to exploding loss.
Experiments were conducted using a private infrastructure, which has a carbon efficiency of 0.432 kgCO$_2$eq/kWh. A cumulative of 2,304 hours of computation was performed, and total emissions are estimated to be 298.6 kgCO$_2$eq.

The fine-tuning of \lscit{}, \lmcit{} and \mcit{} on CHANGE-it took 64 GPU hours each on A100 64Gb GPUs. With the costly synthetic data generation, all together resulted in approximately 2000 GPU hours. Experiments were conducted using the LEONARDO cluster, which has a carbon efficiency of 0.432 kgCO$_2$eq/kWh. A cumulative of 2000 hours of computation was performed on hardware similar to A100 PCIe 40/80GB (TDP of 250W).
Total emissions are estimated to be 216 kgCO$_2$eq \cite{luccioni2019quantifying}.

Thus, we estimate that the total emissions for experiments in this study amount to approximately 515 kgCO$_2$eq.

\end{document}